\documentclass{article}
\usepackage{iclr2027_conference,times}
\usepackage[utf8]{inputenc}
\usepackage[T1]{fontenc}
\usepackage{amsmath,amssymb}
\usepackage{booktabs}
\usepackage{tabularx}
\usepackage{graphicx}
\usepackage{float}
\usepackage{placeins}
\usepackage{url}
\usepackage{xcolor}
\definecolor{ReviewPurple}{RGB}{0,0,0}
\definecolor{Revision}{RGB}{0,0,0}
\definecolor{NewRevision}{RGB}{0,0,0}
\newenvironment{revision}{\begingroup\color{Revision}}{\endgroup}
\newcommand{\rev}[1]{\textcolor{Revision}{#1}}
\newcommand{\newrev}[1]{\textcolor{NewRevision}{#1}}
\usepackage[hidelinks]{hyperref}
\iclrfinalcopy
\title{Selecting Repetition Counts Across Model Scales in Data-Constrained Pretraining}
\author{Ziyue WANG \qquad T. Kanamori\\
Institute of Science Tokyo\\
\texttt{wang.z.7a93@m.isct.ac.jp}}
\hypersetup{
  pdftitle={Selecting Repetition Counts Across Model Scales in Data-Constrained Pretraining},
  pdfauthor={Ziyue WANG; T. Kanamori}
}
\begin{document}
\maketitle
\fancyhead{}
\renewcommand{\headrulewidth}{0pt}
\suppressfloats[t] 
\begin{abstract}
{\color{ReviewPurple}
The repetition count that works best for a small language model may not remain best at a larger scale. We study this effect in pretraining with a finite target corpus mixed with generic data at a fixed target fraction. On Wikipedia-derived data and Proof-Pile-2, the ranking of measured repetition counts changes with model size, and a 520M Proof-Pile-2 experiment confirms that reducing repetition from sixteen to eight improves loss while using fewer training tokens. \newrev{We use loss curves from several smaller models to retain a short list of promising repetition counts for evaluation at a larger scale.} On PubMed and Caselaw, candidate sets fixed before target-model training retain the lowest-loss measured count on the original evaluation grids at both 200M and 520M. This supports candidate retention as a practical alternative to exact point prediction. We also relate the pruning regression to an empirical scaling model with two opposing repetition-dependent loss terms. A first-order expansion in log model size yields the linear form used by the selection rule, providing a scaling-based interpretation of the candidate-selection procedure.
}

\end{abstract}
\section{\textcolor{ReviewPurple}{Introduction}}

\begin{figure}[t]
    \centering
    \includegraphics[width=\textwidth]{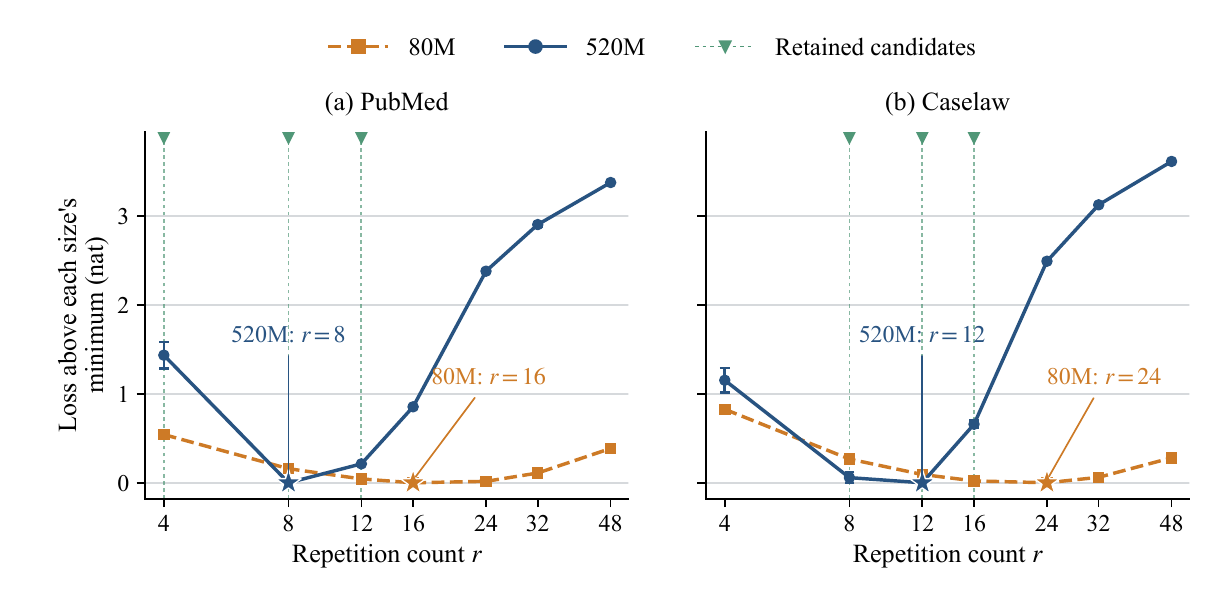}
    \caption{\textcolor{Revision}{\textbf{Repetition preferences change with model size.}
    Here $r$ counts passes over a fixed 100M-token target corpus; M denotes
    millions of model parameters. The vertical axis is held-out target
    cross-entropy in nats minus each size's lowest measured mean.
    Stars mark these minima. Separate markers identify the three retained
    candidates selected from 35M--80M curves before 520M training. Bars show across-seed standard
    deviations (80M: three seeds; 520M: seven). Smaller-model losses use
    a selection split; 520M uses a separate test split. Lines connect
    measurements. Section~\ref{sec:prospective} evaluates the candidates.}}
    \label{fig:headline}
\end{figure}

{\color{ReviewPurple}
Large language model training requires choices about model size, data,
and training duration. Exhaustive comparisons at the intended size are
costly. Scaling laws make smaller experiments useful for these decisions
by relating loss to model size and training tokens
\citep{kaplan2020scalinglawsneurallanguage,hoffmann2022trainingcomputeoptimallargelanguage}.
Related methods use smaller models to select data mixtures
\citep{liu2025regmixdatamixtureregression}.

When valuable data are scarce, another choice is how often to repeat them.
Reusing a token need not have the same value as adding a new one.
Repetition-aware scaling laws account for finite data and mixtures of
scarce and abundant sources
\citep{muennighoff2025scalingdataconstrainedlanguagemodels,sedova2026scalinglawsmixturepretraining,lovelace2026prescriptivescalinglawsdata}.
Their predictions depend on the training design; a repetition count
preferred by a smaller model need not remain best at a larger size.

We study a finite \emph{target corpus} mixed with a larger \emph{generic
corpus}. The repetition count $r$ is the nominal number of passes over the
target pool. We hold the target fraction fixed, so increasing $r$ adds
generic tokens and lengthens training as well as repeating the target.
Each configuration completes its own training schedule. We ask how these
configurations change rank with model size and whether smaller-model
curves can guide their selection.

The initial Wikipedia-derived and Proof-Pile-2 experiments show reversals
between measured counts. Two independent experiments at 520M confirm that
eight repetitions outperform sixteen on Proof-Pile-2 while using half
the training tokens. Four repetitions give higher loss than eight.
These results establish the two pairwise comparisons and motivate a
search for a low-loss region.

We use curves from several smaller sizes to retain three candidates
before larger-model training. Full sweeps on PubMed and Caselaw show that
both sets retain the lowest measured mean at 520M
(Figure~\ref{fig:headline}). A later 200M experiment fixes point predictions
and retained sets before training. It separately tests whether the
predicted count is useful and whether the set retains a good alternative.
Finer measurements resolve the valleys around the original grid minima.

An empirical two-effect model describes how the balance of decreasing
and increasing loss terms changes with size. Its extension estimates
minimizing counts, low-loss intervals, and the extra loss, or \emph{regret},
from nearby choices. We assess smaller-model predictions at an unseen model size and,
separately, fit the empirical extension to predict counts excluded from
fitting.

Our contributions are:
\begin{itemize}
    \item \textbf{Ranking reversals under completed training schedules.}
    Experiments quantify the loss incurred by reusing smaller-model choices.
    \item \textbf{Candidate pruning with larger-model validation.}
    Smaller-model curves select useful three-count sets at 200M and 520M.
    \item \textbf{Empirical estimates of minima and selection regret.}
    We derive minimizing counts and low-loss intervals, and test curve
    predictions with new model-size and repetition-count measurements.
    The same model also gives a local scaling-based interpretation of the
    pruning regression.
\end{itemize}
}

\section{\textcolor{ReviewPurple}{Related Work}}\label{sec:related}

{\color{ReviewPurple}
\paragraph{Scaling model size, data, and compute.}
Empirical scaling laws relate loss to parameters, training tokens, and
compute \citep{kaplan2020scalinglawsneurallanguage}.
Hoffmann et al. show that revised estimates change the preferred
allocation between parameters and tokens
\citep{hoffmann2022trainingcomputeoptimallargelanguage}.
Later work studies deviations from simple power laws and extrapolation
\citep{caballero2023brokenneuralscalinglaws}.
We examine the resulting training decision when the target data must be
repeated.

\paragraph{Selecting data with smaller experiments.}
RegMix uses small proxy models to predict mixture weights for a larger
run \citep{liu2025regmixdatamixtureregression}.
DataDecide and AutoScale examine corpus rankings or domain weights that
change with scale
\citep{magnusson2025datadecidepredictbestpretraining,kang2025autoscalescaleawaredatamixing}.
Related mixture laws and multi-fidelity methods combine experiments at
different sizes or costs to guide data allocation
\citep{que2024dcptlaw,ye2025datamixinglawsoptimizing,ge2024bimix,shukor2025scalinglawsoptimaldata,yen2025datamixtureoptimization}.
Methods such as $\mu$P transfer model parameterizations and optimization
hyperparameters \citep{yang2022tensorprogramsvtuning}.
Our decision is the repetition count, which changes both token allocation
and the completed schedule at a fixed mixture fraction.

\paragraph{Scaling with repeated data.}
Repeated exposures differ from fresh data
\citep{hernandez2022scalinglawsinterpretabilitylearning,xue2023repeatnottorepeat}.
Data-constrained laws track unique data and epochs or discount repetition
through effective token counts
\citep{muennighoff2025scalingdataconstrainedlanguagemodels,xu2026dataconstrainedlanguagemodelpretraining}.
Sedova et al. observe model-size-dependent repetition preferences,
including at fixed mixture fractions, and estimate repetition-aware
optima from intermediate states of shared runs
\citep{sedova2026scalinglawsmixturepretraining}.
Lovelace et al.'s decreasing data term and scale-dependent repetition
penalty also permit valleys and ranking changes
\citep{lovelace2026prescriptivescalinglawsdata}.
Zhou et al. match repetition across proxy and target token horizons
\citep{zhou2026repetitionmismatchdatamixture}; InfoLaw studies efficient
configurations under constrained data
\citep{liu2026infolawinformationscalinglaws}.
A concurrent study finds mildly increasing preferred repetition with
model size under a fixed-tokens-per-parameter design, illustrating the
importance of the training budget
\citep{li2026scalingdomaindatarepetition}.
Our experiments compare independently completed schedules and test
candidate selection at larger sizes. Appendix~\ref{app:H} details the
design differences; Section~\ref{sec:additional-counts} compares empirical
curve predictions with the adapted Lovelace form.
}

\section{\rev{Experimental Setup and Evaluation}}\label{sec:method}

\subsection{Training configurations}
The target pool contains $S_t=100$M tokens and contributes a fraction
$h=0.8$ of training tokens. A larger FineWeb-derived generic corpus supplies
the rest. The nominal repetition count $r$ determines total training tokens:
\begin{equation}
    D_{\mathrm{total}}(r)=\frac{rS_t}{h}=125\mathrm{M}\,r.
    \label{eq:total-tokens}
\end{equation}
We train each combination of $r$ and seed independently to completion,
with a cosine learning-rate schedule over its prescribed token count.
The losses are endpoints of separate runs, not checkpoints of one longer
run. Increasing $r$ changes target exposure, generic tokens, and schedule
length together.

\begin{revision}
\subsection{Loss and evaluation measures}
Let $\mathcal R_N$ be the tested counts at model size $N$ and $L_N(r)$
the mean final held-out target cross-entropy across seeds, in nats.
The lowest-loss count on this measured grid is
\begin{equation}
    r_N^{\mathrm{obs}}\in\arg\min_{r\in\mathcal R_N}L_N(r).
    \label{eq:candidate-minimizer}
\end{equation}
This observed choice differs from a predicted discrete choice
$\widehat r_{N_t}$ and a continuous model minimum $r_*(N)$.
Define excess loss as $E_N(r)=L_N(r)-\min_{u\in\mathcal R_N}L_N(u)$.
A retained set $S\subseteq\mathcal R_{N_t}$ incurs loss gap
$\min_{r\in S}E_{N_t}(r)$ at the larger size $N_t$; a single evaluated
choice incurs $E_{N_t}(\widehat r_{N_t})$.
Our direct-reuse baseline transfers the smaller-model choice\footnote{When
that count is absent from the larger-model grid, we evaluate the closest
tested count in $\log r$, breaking ties toward smaller $r$.}
\citep{liu2025regmixdatamixtureregression,magnusson2025datadecidepredictbestpretraining}.
Complete larger-model sweeps evaluate both this baseline and retained sets.
Section~\ref{sec:scale-model} describes how smaller-model losses select
candidate sets or individual counts; Appendix~\ref{app:I} gives the
full training settings, grids, and seed allocation.
\end{revision}

\section{\textcolor{ReviewPurple}{Ranking Changes Across Model Sizes}}\label{sec:80m}

\subsection{\rev{Ranking reversals at smaller sizes}}\label{sec:rankings}

\begin{figure}[!htb]
    \centering
    \includegraphics[width=\textwidth]{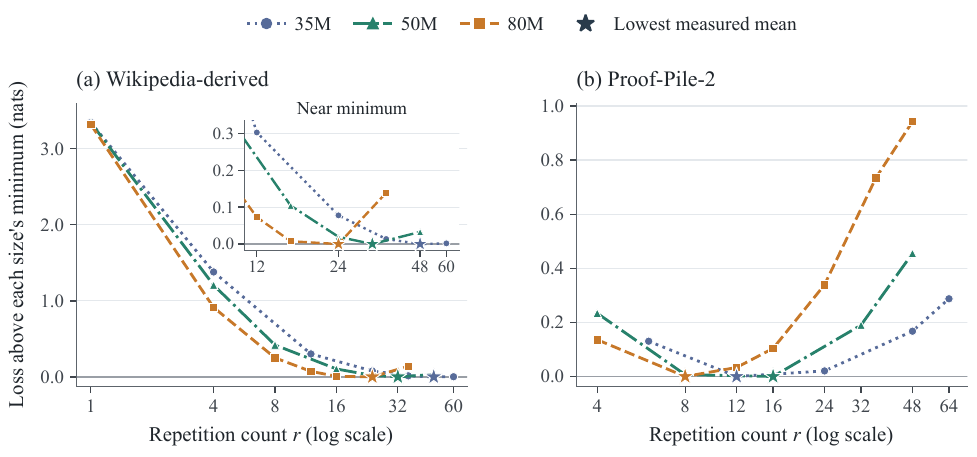}
    \caption{\textcolor{ReviewPurple}{Measured loss versus repetition count
    $r$ at 35M, 50M, and 80M. Each curve subtracts its own lowest mean
    held-out target loss, in nats. Markers show tested counts on the
    original size-dependent grids; stars mark the measured minima.
    The Wikipedia inset enlarges the shallow minima. Lines connect
    measurements. Appendices~\ref{app:D} and~\ref{app:J} report losses
    and variation across seeds.}}
    \label{fig:transfer}
\end{figure}

{\color{ReviewPurple}
\rev{The initial Wikipedia-derived and Proof-Pile-2 experiments use 35M, 50M,
and 80M models.} Figure~\ref{fig:transfer} compares the measured curves at 35M, 50M, and
80M. Because grids differ across sizes, we establish ranking reversals
using pairs measured at both sizes.
}

{\color{ReviewPurple}
For Wikipedia-derived data, $r=36$ has lower mean loss than $r=24$ at 35M.
At 80M, the ordering reverses: 24 has lower loss by 0.139 nat.
The 35M winner, 48, was not tested at 80M; the direct-reuse mapping
therefore uses 36.
The 80M minimum is shallow: 16 is only 0.008 nat worse than 24.
Appendix~\ref{app:D} gives the full curves and shows that the initial
benefit of repetition also appears on clean held-out examples.

For Proof-Pile-2, $r=16$ has lower mean loss than $r=8$ at 50M.
At 80M, 8 gives the lowest measured loss. Reusing 16 raises loss by
0.104 nat, and all three paired differences favor 8.
Extrapolating the smaller models' selected counts also misses 8,
even after adding neighboring candidates (Appendix~\ref{app:J}).
The pruning rule instead uses the full loss curves.

These completed-run comparisons show that repetition rankings can
reverse as model size grows.
}

\begin{revision}
Let $N_s<N_t$ be two model sizes and $a<b$ two counts evaluated at both.
Suppose $b$ has lower loss at the smaller size. Its advantage and the
change in each count's loss are
\begin{align}
 m_s&=L_{N_s}(a)-L_{N_s}(b)\geq0,\label{eq:small-margin}\\
 \delta(r)&=L_{N_t}(r)-L_{N_s}(r).\label{eq:scale-change}
\end{align}
The improvement from choosing $a$ instead of $b$ at the larger size is exactly
\begin{equation}
 G_{b\to a}(N_t)=L_{N_t}(b)-L_{N_t}(a)
              =\delta(b)-\delta(a)-m_s.
 \label{eq:rank-decomposition}
\end{equation}
The ranking reverses when the relative change $\delta(b)-\delta(a)$
exceeds the smaller-model advantage (Table~\ref{tab:rank-decomposition}).
A more precise smaller-model curve estimates $m_s$ better but does not
determine this change. Appendix~\ref{app:K} gives the finite-set results.
\end{revision}

\begin{revision}
\subsection{Confirmation at 520M}\label{sec:520m}
The 80M result motivated a 520M comparison of $r\in\{4,8,16\}$,
fixed before evaluating 520M losses. Changing from 16 to 8 lowers
loss by 0.524 nat across six paired seeds. An independent experiment
with six new seeds gives a 0.509-nat reduction; all twelve differences
favor 8 (Figure~\ref{fig:520m}). This choice also halves the token
budget, from 2B to 1B. Reducing the count to 4 raises loss in every
seed of the first experiment, locating a measured valley around 8.
Appendix~\ref{app:L} gives seed-level results and uncertainty.

\begin{figure}[htb]
\color{Revision}
\centering
\includegraphics[width=\textwidth]{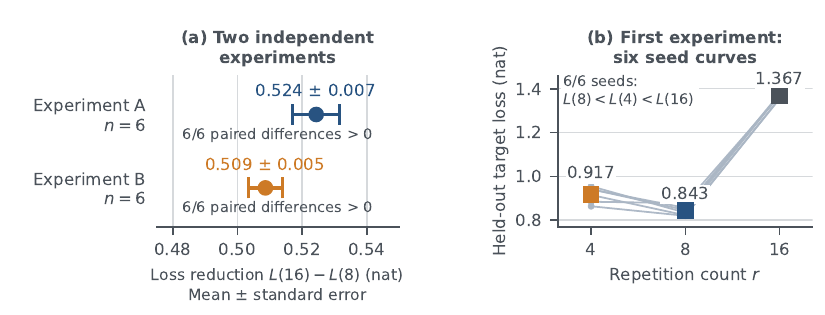}
\caption{Proof-Pile-2 at 520M. Left: mean loss reduction from 16 to 8, with standard-error bars,
for two independent six-seed experiments. All six paired differences
are positive in each experiment. Right:
six seed curves from the first experiment, all satisfying
$L(8)<L(4)<L(16)$; squares and labels give mean losses.}
\label{fig:520m}
\end{figure}
\end{revision}

\begin{revision}
\section{Loss Models and Repetition Selection}\label{sec:scale-model}

\subsection{A model of repetition-dependent loss}\label{sec:two-effects}
Further training can lower loss through additional generic data and
optimization, while repeatedly exposing a fixed target pool can raise
held-out target loss. Motivated by repetition-aware scaling laws
\citep{sedova2026scalinglawsmixturepretraining,lovelace2026prescriptivescalinglawsdata},
we model these directions over a limited range of sizes and counts as
\begin{equation}
 L_N(r)=A(N)+N^\eta U(r)+N^{\eta+q}V(r),
 \qquad U'(r)<0,\quad V'(r)>0,\quad q>0.
 \label{eq:two-effect-model}
\end{equation}
Here $A(N)$ is common to all counts, $\eta$ controls overall amplitude,
and $q>0$ makes the increasing term grow relative to the decreasing one.
\newrev{The two terms provide an empirical description of how loss varies with repetition.}
For $a<b$, write $C=U(a)-U(b)>0$ and $D=V(b)-V(a)>0$. Then
\begin{equation}
 G_{b\to a}(N)=N^\eta(DN^q-C).
 \label{eq:pairwise-crossing}
\end{equation}
Since $DN^q-C$ increases with $N$, a fixed pair can reverse order at
most once within the range where the model holds.
\newrev{For the parametric forms tested in Appendix~\ref{app:q-sign-check}, fitting
$q$ without restricting its sign gives positive estimates on all four corpora.
For PubMed and Caselaw, the seed bootstrap percentile intervals also lie above zero.}

\newrev{At fixed unique-data size, the adapted Lovelace form is the $\eta=0$
case of this model (Appendix~\ref{app:lovelace-comparison}). Freeing $\eta$
allows both terms to vary in overall amplitude with size;
Section~\ref{sec:additional-counts} evaluates the resulting improvement
in held-out loss prediction.}

\subsection{Candidate selection by local regression}\label{sec:candidate-pruning}
On a common finite grid, suppose a minimizing count $r_0$ stays fixed
near model size $N_0$. Under \newrev{Equation~\eqref{eq:two-effect-model}}, subtracting
$L_N(r_0)$ removes $A(N)$. With $x=\log(N/N_0)$,
\[
 E_N(r)=C_re^{\eta x}+D_re^{(\eta+q)x}
       =c_r+d_rx+O(x^2),
\]
where $C_r=N_0^\eta[U(r)-U(r_0)]$, $D_r=N_0^{\eta+q}[V(r)-V(r_0)]$,
$c_r=C_r+D_r$, and $d_r=\eta C_r+(\eta+q)D_r$.
The first-order term is affine in $\log N$, motivating a separate
regression for each count. The observed source minima change across
35M--80M, so a single fixed-minimizer neighborhood does not cover that
range. \newrev{Appendix~\ref{app:local-pruning} details this local interpretation.
We evaluate extrapolation to larger model sizes in Section~\ref{sec:prospective}.}

The candidate rule fits the regressions directly on a common count grid
at 35M, 50M, and 80M. At each size, we resample whole seed curves, average
them, and subtract the resulting curve's minimum. For each resample and
count, we regress this excess loss on $\log N$, predict it at the target
size, and clip negative predictions to zero \emph{before} averaging
across resamples. We round scores to $10^{-9}$ and retain the three
lowest-scoring counts, breaking ties toward smaller $r$.
\newrev{The expansion motivates the regression form. Bootstrap resampling and clipping
are additional steps in the selection procedure. Retaining three candidates
provides alternatives to a single prediction while limiting the number of
configurations evaluated at the target size. On PubMed and Caselaw, the retained
sets contain the count with the lowest measured mean loss on the original
seven-count grids at both 200M and 520M (Section~\ref{sec:prospective}).}
Appendix~\ref{app:pruning-procedure} gives the full rule.

\subsection{Point selection by loss minimization}\label{sec:simple-extrapolation}
Our simple specialization fixes $\eta=0$, $q=1$, with a decreasing
$r^{-1/2}$ term and an increasing term proportional to $Nr$:
\begin{equation}
    \widehat L_N(r)=A_N+B_{\mathrm{dec}} r^{-1/2}
       +P_{\mathrm{inc}}\left(\frac{N}{N_{50}}\right)r.
    \label{eq:app-posthoc-fit}
\end{equation}
\newrev{Fixing $q=1$ makes the increasing term linear in model size and avoids
estimating an additional exponent. In the 200M experiments, fixing $q=1$
preserves the original-grid choices and predicts nearby loss differences
more accurately than fitting $q$ freely
(Appendix~\ref{app:200m-frozen-predictions}, Table~\ref{tab:app200-frozen-functions}).}
Here $N_{50}$ is the 50M model size under the parameter-count convention
being used. \newrev{For each corpus, we first average the loss across seeds at each
source model size and repetition count. We then fit Equation~\eqref{eq:app-posthoc-fit}
to these mean losses by least squares, giving equal weight to each combination
of model size and repetition count. Each source model size has its own offset
$A_N$, while $B_{\mathrm{dec}}$ and $P_{\mathrm{inc}}$ are shared across sizes.
The exponents are fixed.} The 200M prediction fits constrain the shared coefficients
to be nonnegative.
We then select $\widehat r_{N_t}\in\arg\min_{r\in\mathcal R_{N_t}}
\widehat L_{N_t}(r)$ on the target candidate grid.
Count selection and predicted loss differences are independent of the
unknown target offset; absolute-loss prediction requires estimating that
offset. \newrev{Appendix~\ref{app:posthoc-controls} gives the fitting details.
Section~\ref{sec:point-evaluation} evaluates the repetition counts and loss
differences predicted by Equation~\eqref{eq:app-posthoc-fit}.
The prediction functions and the choices on the original seven-count grids
were fixed before 200M training. The 80M and 520M checks were conducted after
the target results were available.}

\subsection{Loss curves and losses near the minimum}\label{sec:structural-regret}
For completed curves, we also fit a richer model with saturation in
size dependence and loss amplitude. Set $n=N/(50\mathrm M)$ and $x=r/8$:
\begin{align}
 w(n)&=\frac{n^q}{1+(n/n_s)^q},\qquad
 F_N(r)=n^\eta[Bx^{-\alpha}+Pw(n)x^\nu],\nonumber\\
 \widehat L_N(r)&=A_N+\frac{KF_N(r)}{K+F_N(r)}.
 \label{eq:saturated-two-effect}
\end{align}
The positive coefficients $B,P$ weight the repetition powers
$\alpha,\nu>0$; $q,n_s>0$ govern size saturation, $K>0$ loss saturation,
and $\eta$ the common amplitude. Each size has its own offset $A_N$.
At a fixed model size, the monotone saturation leaves the minimum
unchanged. Differentiating $F_N$ with respect to $\log r$ and setting
the derivative to zero balances the decreasing and increasing terms,
giving the unique model minimum:
\begin{equation}
 r_*(N)=8\left[\frac{\alpha B}{\nu P}
       (n^{-q}+n_s^{-q})\right]^{1/(\alpha+\nu)}.
 \label{eq:structural-valley}
\end{equation}
Model regret is the extra predicted loss
$\widehat R_N(r)=\widehat L_N(r)-\widehat L_N(r_*)$.
A second-order Taylor expansion in $\log(r/r_*)$ has no linear term
because the derivative vanishes at the minimum. Subtracting the minimum
loss removes the constant, leaving a positive quadratic term:
\begin{equation}
 \widehat R_N(r)=\frac{k_N}{2}[\log(r/r_*)]^2
                 +O(|\log(r/r_*)|^3),\qquad k_N>0.
 \label{eq:local-regret}
\end{equation}
Here $k_N$ is curvature in log repetition. The quadratic local cost
motivates finer evaluation near $r_*$; a chosen regret threshold defines
a model-estimated low-loss interval. Appendix~\ref{app:structural-regret}
derives the exact expression. The additive simple model is a limiting
case and has the same quadratic local form.
\newrev{Section~\ref{sec:additional-counts} evaluates the minima and loss predictions
obtained by fitting Equation~\eqref{eq:saturated-two-effect} to the seven
original counts at 200M (Figure~\ref{fig:intermediate-coarse-fine}).
Appendix~\ref{app:structural-fitting-checks} reports numerical estimates of the
minimum location, curvature, and model regret at 520M. For the simple additive
model, Section~\ref{sec:point-evaluation} and Appendix~\ref{app:200m-local-predictions}
compare predicted loss differences and low-loss intervals with the 200M measurements.}
\end{revision}

\begin{revision}
\section{Evaluating Repetition Selection and Loss Prediction}\label{sec:larger-evaluation}

PubMed and Caselaw use three seeds at each smaller size and seven at 520M.
Candidate sets are fixed from 35M/50M/80M selection losses before
520M training and evaluated on a separate test split. The later 200M
study fixes predictions and sets before training, then uses three paired
seeds per count on the same test split. Both studies use the original
grid $\{4,8,12,16,24,32,48\}$. At 200M, three preplanned counts
$\{6,10,14\}$ and two later exploratory counts $\{9,11\}$ yield a
twelve-count grid (Appendix~\ref{app:intermediate-validation}).

\subsection{Candidate retention at larger sizes}\label{sec:prospective}
Before 520M training, the rule in Section~\ref{sec:candidate-pruning}
selects $\{4,8,12\}$ for PubMed and $\{8,12,16\}$ for Caselaw.
Both sets contain the lowest measured mean among seven tested counts,
at 8 and 12, and span observed valleys (Figure~\ref{fig:headline}).
Their best counts outperform reuse of the 80M winners by 0.854 and
2.492 nat. Across 262 retrospective checks that vary smaller-model
sizes and subsets of existing seeds, the rule also retains the lowest
measured loss on the original 80M or 520M grids
(Appendix~\ref{app:pruning-source-sensitivity}).

At 200M, the frozen sets are $\{8,12,16\}$ and $\{12,16,24\}$.
Count 12 has the lowest loss in all three seeds on each original
seven-count grid, so both sets retain it. On the refined twelve-count
grid, the lowest means are at 11 and 12; the original sets incur
0.006 and 0 nat.

\paragraph{Baselines and cost.}
\newrev{As a baseline, we retain the three counts with the lowest measured mean loss at
80M: $\{12,16,24\}$ for PubMed and $\{16,24,32\}$ for Caselaw. At 520M, our
retained sets achieve lower minimum measured mean losses than these baseline sets.
The reductions are 0.213 nat on PubMed and 0.658 nat on Caselaw
(Appendix~\ref{app:pruning-520m-results}).}

\newrev{For the original 520M study, we estimate the cost of applying our selection rule,
including all runs at smaller sizes and only the retained counts at 520M. The
combined $6ND$ cost across both corpora would be 65.6\% lower than running all
seven counts at 520M. The validation experiments trained all seven counts
(Appendix~\ref{app:pruning-cost}). Appendix~\ref{app:pruning-statistical} reports
the statistical analysis of the retained sets.}

\subsection{Point predictions and nearby losses}\label{sec:point-evaluation}
\label{sec:intermediate-validation}
The simple model in Section~\ref{sec:simple-extrapolation}, fitted post hoc
to 35M/50M curves, selects the measured Wikipedia and Proof-Pile-2 80M
minima (Appendix~\ref{app:posthoc-controls}). Its Proof-Pile-2 extrapolation
to 520M selects 4 instead of the better measured count 8
(Appendix~\ref{app:held-scale-check}).
\newrev{For the 200M test, we fit Equation~\eqref{eq:app-posthoc-fit} to the 35M, 50M, and
80M loss curves. We fix the prediction functions and the choices on the original
seven-count grids before 200M training. The continuous minima are 9.659 for PubMed
and 11.431 for Caselaw. Minimizing the predicted loss on the original grids
selects $r=8$ and $r=12$, respectively. Relative to the lowest measured mean loss
on each original grid, these choices incur excess losses of 0.038 nat for PubMed
and 0 nat for Caselaw.}

\newrev{After extending the evaluation grid from seven to twelve counts, every seed curve
decreases and then increases. We then apply the same prediction functions to
integer counts without refitting their parameters. This selects $r=10$ for PubMed
and $r=11$ for Caselaw. Their measured mean losses are 0.003 and 0.008 nat above
the lowest measured means on the respective twelve-count grids
(Table~\ref{tab:intermediate-predictions}).
Appendix~\ref{app:200m-frozen-predictions} reports the other fits fixed before
200M training.}

\begin{table}[htb]
\color{Revision}
\centering\normalsize
\caption{200M point choices and their measured excess loss in nats.
The seven-count choices were fixed before training. Integer choices
apply the unchanged prediction functions after refinement.
Zero denotes selection of the lowest measured mean on the stated grid.}
\label{tab:intermediate-predictions}
\begin{tabular}{lrrrr}
\toprule
&\multicolumn{2}{c}{PubMed}&\multicolumn{2}{c}{Caselaw}\\
Prediction and evaluation grid & $r$ & Excess loss & $r$ & Excess loss\\
\midrule
Original choice, seven counts & 8 & 0.038 & 12 & 0\\
Original choice, twelve counts & 8 & 0.044 & 12 & 0\\
Same function on integer $r$, twelve counts & 10 & 0.003 & 11 & 0.008\\
\bottomrule
\end{tabular}
\end{table}

\begin{figure}[htb]
\color{Revision}
\centering
\includegraphics[width=\textwidth]{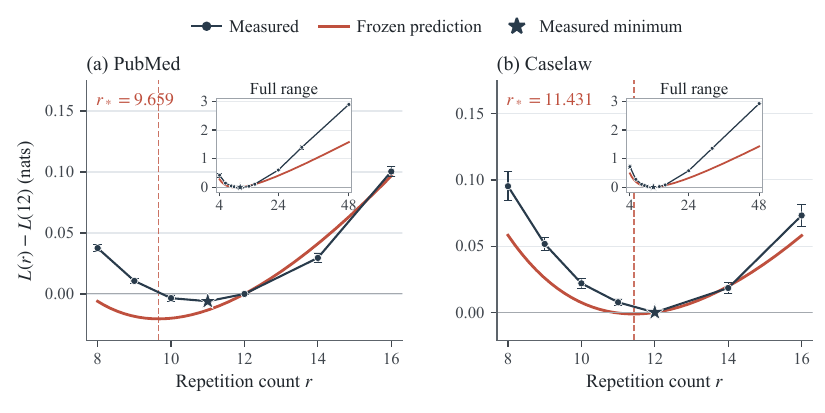}
\caption{Predictions fixed from 35M/50M/80M and measured 200M loss
versus repetition count $r$, each relative to its value at 12.
No 200M parameters are fitted. Main panels show the valley; insets show
the full range. Bars are sample SDs of three paired seed differences.
Dashed lines mark predicted continuous minima, stars measured minima.}
\label{fig:intermediate-predictions}
\end{figure}

\newrev{Figure~\ref{fig:intermediate-predictions} compares predicted and measured loss
differences $L(r)-L(12)$ at 200M. The predictions use
Equation~\eqref{eq:app-posthoc-fit} fitted only to the smaller-model data. On the
seven measured counts $\{8,9,10,11,12,14,16\}$, the RMSE is 0.021 nat for PubMed
and 0.019 nat for Caselaw. The measured second difference $L(10)-2L(11)+L(12)$ is
positive for all six seed curves, consistent with local convexity near the
minimum. Prediction errors increase at higher repetition counts. The boundaries of
predicted intervals that keep loss within a specified tolerance of the minimum are
less accurate than the predicted minimum locations.
Appendix~\ref{app:intermediate-model-checks} reports these comparisons in detail.}

\end{revision}

\subsection{\rev{Predicting losses at additional repetition counts}}\label{sec:additional-counts}
\begin{figure}[htb]
\color{ReviewPurple}
\centering
\includegraphics[width=\textwidth]{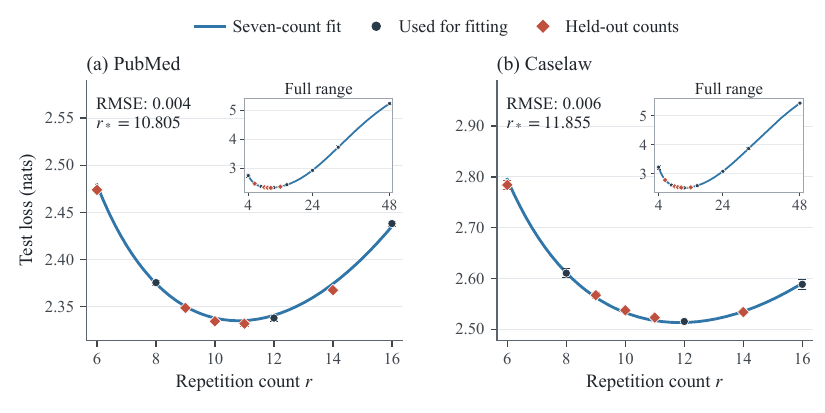}
\caption{New 200M measurements predicted from seven counts.
\newrev{Equation~\eqref{eq:saturated-two-effect}}, including its offset, is fitted
only at $r\in\{4,8,12,16,24,32,48\}$. Diamonds mark the five
excluded counts $\{6,9,10,11,14\}$ used for RMSE. Main panels show
the valley and insets the full range. Loss is absolute test loss in
nats; bars are sample SDs across three seeds.}
\label{fig:intermediate-coarse-fine}
\end{figure}

\newrev{Using the completed 200M measurements, we fit the fixed-size form of
Equation~\eqref{eq:saturated-two-effect} to the seven original counts for each
corpus. The five additional counts are excluded from both shape and offset
estimation. Predictions for these five counts have RMSEs of 0.004 nat for PubMed
and 0.006 nat for Caselaw (Figure~\ref{fig:intermediate-coarse-fine}).}
Its continuous minima, 10.805 and 11.855, select the measured best
counts, 11 and 12. \newrev{Accurate predictions at the five held-out counts support the fitted repetition-dependent curve shape at 200M.}

\begin{samepage}
\newrev{In a separate analysis, we compare three loss models by fitting each jointly
to the original seven counts at 35M, 50M, 80M, 200M, and 520M. We reevaluate
the losses of the smaller models on the common test split. Each loss model
shares its shape parameters across sizes and has a separate offset at each size.
Using the same seven repetition counts at each size, allowing $\eta$ to vary
in the adapted Lovelace model improves loss predictions at the five excluded
200M counts. The model in Equation~\eqref{eq:saturated-two-effect} reduces
these prediction errors further (Table~\ref{tab:new-count-models}).}
\newrev{Freeing $\eta$ reduces held-out loss-prediction errors while preserving
the same repetition choices and measured selection losses. The saturated
extension reduces RMSE further but incurs larger selection losses in these
joint fits.} \newrev{Appendix~\ref{app:intermediate-model-checks} examines the remaining
amplitude errors of the joint cross-size fits.}
\par
\end{samepage}

\begin{table}[htb]
\color{ReviewPurple}
\centering\normalsize
\caption{Matched fits to seven counts at 35M/50M/80M/200M/520M,
including offsets. RMSE uses the five excluded 200M counts. Selection
loss is the measured excess loss of the chosen count over the
twelve-count minimum. All values are in nats.}
\label{tab:new-count-models}
\begin{tabular}{lrrrr}
\toprule
&\multicolumn{2}{c}{New-count RMSE}&\multicolumn{2}{c}{Selection loss}\\
Model&PubMed&Caselaw&PubMed&Caselaw\\
\midrule
\newrev{Adapted Lovelace model, Eq.~\eqref{eq:app-lovelace-shape}}&0.238&0.224&0.003&0.008\\
Same repetition terms, free $\eta$&0.159&0.184&0.003&0.008\\
Saturated extension, \newrev{Eq.~\eqref{eq:saturated-two-effect}}&0.108&0.074&0.006&0.018\\
\bottomrule
\end{tabular}
\end{table}

\FloatBarrier
{\color{ReviewPurple}
\section{Discussion and Conclusion}\label{sec:discussion}
Mixture pretraining with a scarce target corpus and abundant generic
data requires a repetition choice that token-count power laws alone
do not resolve \citep{sedova2026scalinglawsmixturepretraining}.
Our completed-run comparisons show that the preferred count can change
with model size. Candidate pruning retains low-loss choices at 200M
and 520M. The finer 200M grid also supports two distinct prediction tests:
smaller-model fits locate the valley, while separate fits to seven
200M counts predict the additional measurements.
Together, these results provide empirical guidance for repetition
selection when more unique target data are unavailable.

\paragraph{Practical use.}
At a fixed mixture fraction, the empirical models estimate a promising
repetition count and the loss associated with nearby choices. The
separate pruning rule uses smaller-model curves to choose a short list
for evaluation at the intended size. Local loss estimates can guide
subsequent refinement while the generic source continues to supply
fresh tokens.

\paragraph{Scope and limitations.}
The main sweeps fix the target-pool size and mixture fraction and use
one model family and training setup. Broader data budgets, mixtures, and architectures
would test the reach of the fitted relationships. Shared fits retain
amplitude errors, and better curve prediction does not always improve
the selected count.
Increasing repetition also changes generic exposure and schedule
length, so the measured curves describe their joint effect.
}

{\color{ReviewPurple}
\subsection*{AI use statement}

\newrev{We used generative AI tools to edit and revise portions of the manuscript
for clarity and readability and to assist with writing and editing portions
of the code used in this study. The authors reviewed all AI-assisted text
and checked the affected code and its outputs for correctness.
The authors take full responsibility for the final manuscript,
code, analyses, and reported results.}
}

\bibliographystyle{iclr2027_conference}
\bibliography{regimen}
\clearpage
\appendix
\raggedbottom
\numberwithin{figure}{section}
\numberwithin{table}{section}
\setcounter{figure}{0}
\setcounter{table}{0}
\section*{Appendix Roadmap}

Appendix~\ref{app:I} gives the shared training settings, corpus splits, and seed allocation, including the order of decisions in the 200M study. Appendix~\ref{app:measured-curves-controls} reports the original loss curves, both Proof-Pile-2 experiments at 520M, and controls for training exposure and schedule length. Appendix~\ref{app:new-experiments} gives the candidate-selection rule, full 520M and 200M evaluations, sensitivity to smaller-model inputs, and statistical and compute comparisons.

Appendix~\ref{app:K} \rev{gives the observed reversal decomposition, finite-set results, and local approximation in log model size.} Appendix~\ref{app:empirical-models} develops the empirical curve and regret, compares it with the Lovelace form, checks the sign of the scale exponent, and evaluates predictions on the additional 200M counts. Appendix~\ref{app:H} relates this setting to existing scaling laws. Appendix~\ref{app:secondary-diagnostics} collects the optional Wikipedia diagnostics on generic-corpus construction, simple selection baselines, gradient alignment, and text overlap.

\section{Experimental Settings}\label{app:I}

This appendix records the settings needed to interpret and reproduce the reported configurations. Repeating pretraining additionally requires access to the external corpora, tokenized pools, tokenizer assets, and compute described below; the numerical loss tables alone are not a substitute for rerunning the training jobs.

\subsection{Shared model and optimization settings}

The nominal 35M, 50M, and 80M models are decoder-only Transformers with 8, 10, and 12 layers; widths 448, 512, and 640; 8 attention heads; context length 1024; and a 32,768-token vocabulary. Their realized parameter counts are 34,453,888, 48,826,368, and 80,710,400. The nominal 520M model has 30 layers, width 1152, 18 heads, and 517,137,408 realized parameters. Parameter counts include embeddings and biases.

All experiments use AdamW with $\beta=(0.9,0.95)$, learning rate $3\times10^{-4}$, weight decay 0.1, gradient clipping at 1.0, a 1\% warmup, cosine decay to 10\% of the peak rate, bfloat16 (bf16) training, and a global batch of 524,288 tokens. The smaller models use microbatch 64. The Proof-Pile-2 520M comparisons use four GPUs with microbatch 4 and 32 gradient-accumulation steps per worker. The PubMed and Caselaw 520M runs use one GPU with microbatch 4 and 128 accumulation steps. Both configurations preserve the global batch of 524,288 tokens at context length 1024. These memory-driven implementation changes do not alter the data, optimizer values, final token count, or final-loss definition.

\subsection{Wikipedia-derived study}

\begin{table}[H]
\centering
\caption{Configuration of the Wikipedia-derived loss curves.}
\label{tab:wikipedia-configuration}
\normalsize
\begin{tabularx}{\linewidth}{p{0.24\linewidth}X}
\toprule
Item & Setting \\
\midrule
Target pool & 100M English-Wikipedia-derived tokens: two 50M-token subpools, one clean-prose construction and one redundancy-injected construction \\
Generic training corpus & FineWeb-derived training stream; generic tokens fill the fraction $1-h=0.2$ of each main training configuration \\
Tokenizer and context & Fixed 32,768-token byte-level byte-pair encoding (BPE); context length 1024 \\
Target evaluation & Mean over the two clean held-out target splits; each constituent split uses at most 2,097,152 tokens \\
Held-out generic loss & Fixed held-out generic data, used only for comparisons between configurations within this study \\
Candidate sets & 35M: $\{1,4,12,24,36,48,60\}$; 50M: $\{1,4,8,16,24,32,48\}$; 80M: $\{1,4,8,12,16,24,36\}$ \\
Seeds & Four per 35M model/count combination except three at $r=24$; three per 50M and 80M combination \\
Training-token allocation & $D_{\mathrm{total}}=125\mathrm{M}r$ tokens, with a separate cosine schedule for each count \\
\bottomrule
\end{tabularx}
\end{table}

The two target subpools and their held-out splits are constructed before training; the text-overlap comparison separately evaluates clean examples, high-overlap examples, and held-out examples from the original Wikipedia-derived corpus. The main loss $L_N(r)$ is the mean on the two clean held-out splits. Section~\ref{app:D} gives all resulting configuration means, standard deviations, and seed counts.

\subsection{Proof-Pile-2 and 520M studies}

\begin{table}[H]
\centering
\caption{Cross-scale comparability of the Proof-Pile-2 experiments.}
\label{tab:target2-comparability}
\normalsize
\begin{tabularx}{\textwidth}{p{0.20\textwidth}p{0.34\textwidth}X}
\toprule
Item & 35M/50M/80M experiments & 520M experiments \\
\midrule
Target corpus and tokenizer & Fixed 100M-token Proof-Pile-2 formal-mathematics/code target, validation split, and 32,768-token tokenizer & Same corpus, split, and tokenizer \\
Generic training corpus & Fixed FineWeb-derived training stream & Same corpus \\
Target evaluation & Fixed held-out target split; at most 2,097,152 tokens & Same split and maximum \\
Held-out generic loss & Fixed held-out generic data used for within-corpus comparisons & Same evaluation data \\
Candidate sets & 35M: $\{6,12,24,48,64\}$; 50M: $\{4,8,16,32,48\}$; 80M: $\{4,8,12,16,24,36,48\}$ & First experiment: $\{4,8,16\}$; second experiment: $\{8,16\}$ \\
Seeds & Four per 35M model/count combination; three per 50M and 80M combination & Six paired seeds in each nonoverlapping seed set \\
Training-token allocation & $D_{\mathrm{total}}=125\mathrm{M}r$ with a separate schedule for each $r$ & Same definition \\
\bottomrule
\end{tabularx}
\end{table}

The fixed-horizon control in Section~\ref{app:fixed-horizon-controls} is the one exception to the token-allocation rule for complete training configurations: it holds total tokens and the cosine horizon at 4B while changing the target fraction. Section~\ref{app:L} reports every seed-level result for the principal 520M comparison.

\subsection{PubMed and Caselaw data and seed allocation}\label{app:pubmed-caselaw-settings}

The targets are CommonPile PubMed and Caselaw Access Project. Each uses
a 100M-token training pool and separate 5M-token selection and test splits.
The target fraction is $h=0.8$, and the generic training stream and model
family follow Appendix~\ref{app:I}. At 35M, 50M, and 80M we evaluate
$r\in\{4,8,12,16,24,32,48\}$ with three seeds, giving 126 smaller-model runs.
All of these runs finish before candidate selection. We then evaluate
the same grid at 520M with seven seeds on the separate test split, giving
98 larger-model runs and 224 completed runs in total. Calculations of
scale and compute use the realized parameter counts in Appendix~\ref{app:I}.

\subsection{Intermediate-model training, evaluation, and decision order}\label{app:200m-settings}

The intermediate model has 203,897,344 parameters, 18 transformer layers,
hidden width 896, 14 attention heads, and context length 1024. It uses the
same tokenizer, fixed 100M-token target pools, generic stream, and target
fraction $h=0.8$ as the PubMed and Caselaw experiments. The planned total
training tokens are $D=rS_t/h=125\mathrm{M}\,r$. Every count has its own
complete warmup and cosine schedule. Optimizer settings follow
Appendix~\ref{app:I}: AdamW, peak learning rate $3\times10^{-4}$,
$\beta=(0.9,0.95)$, weight decay 0.1, gradient clipping at 1.0,
1\% warmup, and decay to 10\% of the peak rate. The global batch is
524,288 tokens; the 200M runs use two GPUs with a microbatch of 32 sequences
per worker and bfloat16 training.

The original grid was
$G_7=\{4,8,12,16,24,32,48\}$. For each corpus, three seed identifiers
were paired across counts. The main experiment therefore contains
$2\times7\times3=42$ independently completed runs. Before examining any
200M results, we saved the smaller-model fits, their choices on $G_7$,
their continuous minima, and the retained sets. The primary fit and
retention rule used the 35M, 50M, and 80M selection curves; a separate
35M/50M analysis was also fixed in advance. The model families and
retention rule had been developed using earlier experiments, including
520M analyses; the numerical 200M predictions themselves used only these
smaller-model curves.

Three additional counts, $\{6,10,14\}$, were prepared before target
results were available and run as the planned refinement when resources
permitted, adding 18 runs. Counts $\{9,11\}$ were chosen after observing
the main results and added 12 exploratory runs. All 72 runs completed.
The final grid is
$G_{12}=\{4,6,8,9,10,11,12,14,16,24,32,48\}$.
Thus the five added counts were not all chosen prospectively, even when
a subsequently evaluated prediction function had already been frozen.

All 200M target losses use the same held-out test stream, with 4,999,168
evaluated tokens per endpoint. We also reevaluated the existing
35M/50M/80M endpoints on this test stream:
$2\times3\times7\times3=126$ evaluations, without additional training.
These evaluations support comparisons on a common split and the new
empirical fits in Section~\ref{app:intermediate-model-checks}. They do not replace the selection data used to
freeze the predictions. Comparisons across 35M, 50M, 80M, 200M, and 520M
use the common seven-count grid; 200M has three seeds per count and
520M has seven.

\section{Measured Loss Curves and Control Experiments}\label{app:measured-curves-controls}

\subsection{Wikipedia-Derived Loss Curves}\label{app:D}

This section reports the complete 35M, 50M, and 80M loss curves behind the Wikipedia-derived result. Every entry is the final held-out target loss $L_N(r)$ of an independently completed configuration.

\subsubsection{Smaller-model curves}

\begin{table}[H]
\centering
\caption{Completed 35M Wikipedia-derived loss curve. The $r=24$ configuration has three completed seeds; all other repetition counts have four. ``Seed SD'' denotes the standard deviation across random seeds.}
\label{tab:wikipedia-35m}
\normalsize
\begin{tabular}{lrrrrrrr}
\toprule
$r$ & 1 & 4 & 12 & 24 & 36 & 48 & 60 \\
\midrule
Mean $L_{35\mathrm{M}}(r)$ & 6.274 & 4.307 & 3.229 & 3.005 & 2.941 & 2.927 & 2.929 \\
Seed SD & 0.026 & 0.019 & 0.011 & 0.004 & 0.003 & 0.002 & 0.004 \\
Seeds & 4 & 4 & 4 & 3 & 4 & 4 & 4 \\
\bottomrule
\end{tabular}
\end{table}

\begin{table}[H]
\centering
\caption{Completed 50M Wikipedia-derived loss curve. Every repetition count has three seeds.}
\label{tab:wikipedia-50m}
\normalsize
\begin{tabular}{lrrrrrrr}
\toprule
$r$ & 1 & 4 & 8 & 16 & 24 & 32 & 48 \\
\midrule
Mean $L_{50\mathrm{M}}(r)$ & 6.256 & 4.118 & 3.330 & 3.014 & 2.930 & 2.910 & 2.943 \\
Seed SD & 0.036 & 0.018 & 0.005 & 0.004 & 0.004 & 0.002 & 0.002 \\
Seeds & 3 & 3 & 3 & 3 & 3 & 3 & 3 \\
\bottomrule
\end{tabular}
\end{table}

The lowest measured 35M loss is at $r=48$, although $r=60$ is only 0.002 nat higher. At 50M, the lowest measured loss is at $r=32$. Reusing 48 raises loss by 0.033 nat, and the three paired differences are $\{0.035,0.029,0.035\}$ nat. We therefore treat the measured 35M minimum as shallow while regarding the 48-versus-32 ordering at 50M as well resolved.

\subsubsection{The 80M curve}

The evaluated 80M candidate set is $\{1,4,8,12,16,24,36\}$. Counts 32 and 48 are both mapped to 36 by the log-distance rule in Section~\ref{sec:method}. This candidate set and mapping were specified before the 80M losses were evaluated.

\begin{table}[H]
\centering
\caption{{\color{ReviewPurple}
Completed 80M Wikipedia-derived loss curve. Every repetition count has three seeds. Loss differences are calculated before rounding the displayed means.
}}
\label{tab:wikipedia-80m}
\normalsize
\begin{tabular}{lrrrrrrr}
\toprule
$r$ & 1 & 4 & 8 & 12 & 16 & 24 & 36 \\
\midrule
Mean $L_{80\mathrm{M}}(r)$ & 6.225 & 3.818 & 3.155 & 2.977 & 2.912 & 2.905 & 3.044 \\
Seed SD & 0.015 & 0.062 & 0.013 & 0.003 & 0.001 & 0.005 & 0.009 \\
Seeds & 3 & 3 & 3 & 3 & 3 & 3 & 3 \\
\bottomrule
\end{tabular}
\end{table}

The lowest measured mean is at $r=24$. Count 16 is only 0.008 nat higher, so the data also support a shallow near-minimum region around 16--24. In contrast, the mapped smaller-model choice $r=36$ is 0.139 nat worse than $r=24$. A conservative calculation that treats the two sets of seed runs as unpaired gives a standard error of approximately 0.006 nat; the across-seed standard deviations are shown in Table~\ref{tab:wikipedia-80m}.

\subsubsection{Near-duplicate diagnostic}

{\color{ReviewPurple}
The target construction includes a clean held-out split and an auxiliary split with unusually high text overlap with the target training pool. The mean difference between clean loss and loss on high-overlap examples increases from 0.126 nat at $r=1$ to 0.553 nat at $r=48$ for 35M, and from 0.129 to 0.579 nat for 50M. These values use the final evaluations of all four and three seeds, respectively. A separate 50M control trained without target tokens ($h=0$) has a gap of 0.074 nat after 6B tokens (two seeds), using the same total token budget as the $r=48$ configuration. Clean evaluation loss also initially decreases with repetition, so the initial improvement is not confined to high-overlap examples. This diagnostic does not identify the cause of the later increase in target loss.
}

\subsection{Proof-Pile-2 Loss Curves and Smaller-Model Predictions}\label{app:J}

This section reports the completed 35M, 50M, and 80M Proof-Pile-2 loss curves. Values are seed-mean final held-out losses in nats. The 35M curve has four seeds per count; the 50M and 80M curves have three.

\subsubsection{Smaller-model loss curves}

\begin{table}[H]
\centering
\caption{Completed Proof-Pile-2 loss curves at 35M and 50M.}
\label{tab:target2-smaller-curves}
\normalsize
\begin{tabular}{crr@{\qquad}crr}
\toprule
Scale & $r$ & Mean loss & Scale & $r$ & Mean loss \\
\midrule
35M & 6  & 0.913 & 50M & 4  & 1.021 \\
35M & 12 & 0.783 & 50M & 8  & 0.795 \\
35M & 24 & 0.804 & 50M & 16 & 0.788 \\
35M & 48 & 0.950 & 50M & 32 & 0.978 \\
35M & 64 & 1.070 & 50M & 48 & 1.243 \\
\bottomrule
\end{tabular}
\end{table}

The lowest measured counts are $r=12$ at 35M and $r=16$ at 50M. Their change is not monotone toward the later 80M value, which is why we distinguish direct reuse from extrapolation through the two smaller-model selections.

\subsubsection{The 80M loss curve}

\begin{table}[H]
\centering
\caption{Completed Proof-Pile-2 loss curve at 80M. Each count has three seeds; ``Seed SD'' denotes the standard deviation across them.}
\label{tab:target2-80m-curve}
\normalsize
\begin{tabular}{crrr}
\toprule
Scale & $r$ & Mean loss & Seed SD \\
\midrule
80M & 4  & 0.898 & 0.009 \\
80M & 8  & 0.762 & 0.004 \\
80M & 12 & 0.796 & 0.005 \\
80M & 16 & 0.866 & 0.007 \\
80M & 24 & 1.101 & 0.007 \\
80M & 36 & 1.497 & 0.015 \\
80M & 48 & 1.704 & 0.010 \\
\bottomrule
\end{tabular}
\end{table}

The lowest mean is at $r=8$. Directly reusing $r=16$ raises loss by 0.104 nat. The three matched-seed differences are $\{0.101,0.110,0.100\}$ nat, with a paired standard error of 0.003 nat, and all favor $r=8$. This is a measured comparison on the evaluated candidate set; it does not identify the lowest-loss untested count.

\subsubsection{Predictions made from the smaller models}

\begin{table}[H]
\centering
\caption{Predictions produced from the completed smaller-model experiments.}
\label{tab:target2-rule-outputs}
\normalsize
\begin{tabularx}{\textwidth}{p{0.30\textwidth}p{0.24\textwidth}X}
\toprule
Rule & Output at 80M & Interpretation \\
\midrule
Direct reuse of 50M selection & $r=16$ & Reuse the 50M minimum \\
Two-point log-linear extrapolation & 24.22, mapped to $r=24$ & Extrapolate the two selected counts, then use the nearest evaluated count \\
Nearby candidate set & $\{16,24,36\}$ & Extrapolated center and its adjacent evaluated counts \\
\bottomrule
\end{tabularx}
\end{table}

The 24.22 prediction uses the realized parameter counts 34,453,888, 48,826,368, and 80,710,400. Using the nominal 35M, 50M, and 80M labels gives 23.38; both calculations map to the evaluated count $r=24$. The candidate set does not contain the observed 80M minimum $r=8$. The later 520M experiments contain no $r=24$ run and therefore do not extend the $r=8$ versus $r=24$ comparison to that scale.

\subsection{Seed-Level Results at 520M}\label{app:L}

\begin{revision}
\begin{figure}[htb]
\color{Revision}
\centering
\includegraphics[width=0.88\textwidth]{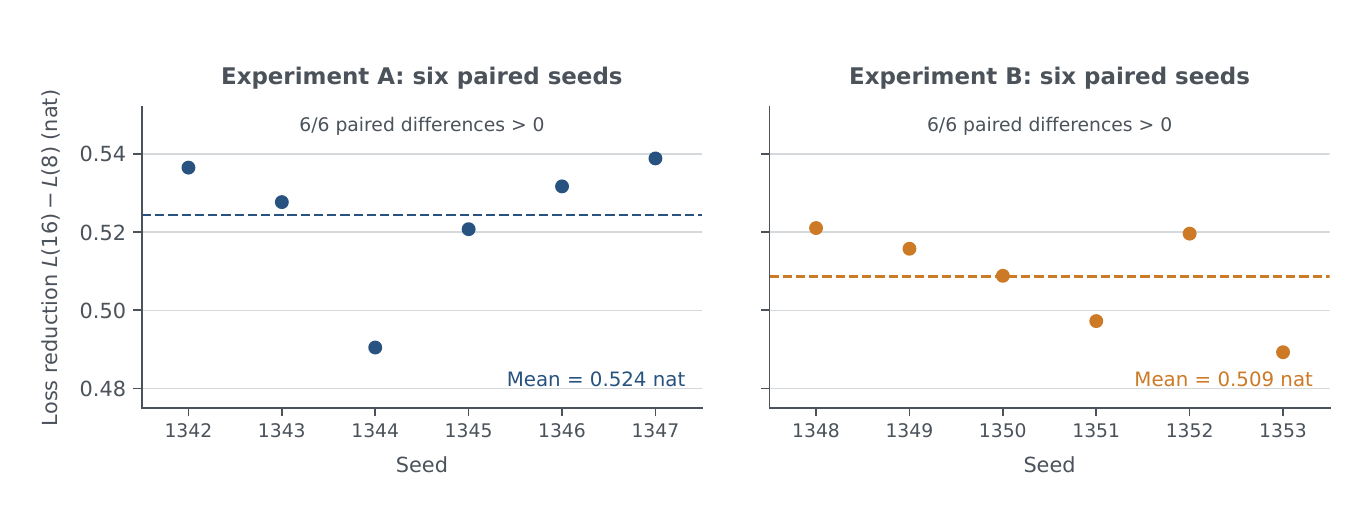}
\caption{Individual paired loss reductions $L(16)-L(8)$ for the two
independent 520M Proof-Pile-2 experiments. The twelve points are grouped
by experiment, with six nonoverlapping seeds each; all are positive.
Dashed lines mark the experiment means.
Figure~\ref{fig:520m} summarizes each experiment with its mean and SE.}
\label{fig:520m-paired-points}
\end{figure}
\end{revision}

This section reports every seed-level result behind Section~\ref{sec:520m}. Losses and paired differences are in nats and are rounded to three decimals. Within this section, $L(r)$ abbreviates $L_{520\mathrm{M}}(r)$.

The observed 80M Proof-Pile-2 comparison determined the 520M pair $r=16$ versus $r=8$. The first 520M protocol specified that pair, six paired seeds, and an additional $r=4$ comparison before any 520M loss was evaluated. The second experiment used six new paired seeds and the same $r=16$ versus $r=8$ contrast. The two seed sets do not overlap and use the same model, corpora, optimizer, and training protocol.

These are complete training configurations rather than compute-matched treatments. At $h=0.8$, $r=4$, 8, and 16 correspond to 0.5B, 1B, and 2B total tokens and use separate cosine-schedule horizons.

\subsubsection{First experiment}

\begin{table}[H]
\centering
\caption{Seed-level results for the first 520M experiment.}
\label{tab:520m-first-seeds}
\normalsize
\begin{tabular}{crrrr}
\toprule
Seed & $L(8)$ & $L(4)$ & $L(16)$ & $L(16)-L(8)$ \\
\midrule
1342 & 0.820 & 0.864 & 1.357 & 0.537 \\
1343 & 0.846 & 0.942 & 1.374 & 0.528 \\
1344 & 0.878 & 0.884 & 1.368 & 0.490 \\
1345 & 0.856 & 0.941 & 1.377 & 0.521 \\
1346 & 0.822 & 0.915 & 1.354 & 0.532 \\
1347 & 0.834 & 0.955 & 1.373 & 0.539 \\
\bottomrule
\end{tabular}
\end{table}

The paired difference has mean 0.524, standard error 0.007, and one-sided 95\% Student-$t$ lower confidence bound 0.510. All six differences favor $r=8$.

\subsubsection{Second experiment with nonoverlapping seeds}

\begin{table}[H]
\centering
\caption{Seed-level results for the second 520M experiment.}
\label{tab:520m-replication-seeds}
\normalsize
\begin{tabular}{crrr}
\toprule
Seed & $L(8)$ & $L(16)$ & $L(16)-L(8)$ \\
\midrule
1348 & 0.852 & 1.373 & 0.521 \\
1349 & 0.848 & 1.364 & 0.516 \\
1350 & 0.854 & 1.362 & 0.509 \\
1351 & 0.853 & 1.350 & 0.497 \\
1352 & 0.846 & 1.366 & 0.520 \\
1353 & 0.880 & 1.369 & 0.489 \\
\bottomrule
\end{tabular}
\end{table}

The mean is 0.509, the standard error is 0.005, and the one-sided 95\% lower confidence bound is 0.498. All six differences again favor $r=8$. Across all twelve paired seeds, the mean is 0.516 and the one-sided lower confidence bound is 0.508; this pooled calculation is a descriptive summary rather than a third experiment.

\subsubsection{Additional \texorpdfstring{$r=4$}{r=4} comparison}

\begin{table}[H]
\centering
\caption{Seed-level contrasts involving $r=4$ in the first experiment.}
\label{tab:520m-r4-seeds}
\normalsize
\begin{tabular}{crrl}
\toprule
Seed & $L(16)-L(4)$ & $L(4)-L(8)$ & Loss order \\
\midrule
1342 & 0.493 & 0.044 & $L(8)<L(4)<L(16)$ \\
1343 & 0.432 & 0.095 & $L(8)<L(4)<L(16)$ \\
1344 & 0.484 & 0.007 & $L(8)<L(4)<L(16)$ \\
1345 & 0.436 & 0.084 & $L(8)<L(4)<L(16)$ \\
1346 & 0.439 & 0.093 & $L(8)<L(4)<L(16)$ \\
1347 & 0.418 & 0.121 & $L(8)<L(4)<L(16)$ \\
\bottomrule
\end{tabular}
\end{table}

The mean losses are 0.843 for $r=8$, 0.917 for $r=4$, and 1.367 for $r=16$. The mean contrasts are 0.450 for $L(16)-L(4)$ and 0.074 for $L(4)-L(8)$. These three counts establish an ordering only among the evaluated candidates; they do not determine the shape of the loss curve between them or a global optimum.

\subsubsection{Pairwise orderings across model size}
Table~\ref{tab:pairwise-crossing} places the 520M comparisons beside the
smaller-model results. One pair changes order between 50M and 80M,
whereas the other does not change order through 520M. The pattern is
consistent with Equation~\ref{eq:pairwise-crossing} within its stated scope.
\begin{table}[htb]
\centering\normalsize
\caption{Proof-Pile-2 differences $G_{b\to a}(N)=L_N(b)-L_N(a)$;
positive values favor the lower count. The second 520M experiment
uses new seeds and evaluates only 8 and 16.}
\label{tab:pairwise-crossing}
\begin{tabular}{lrrrr}
\toprule
Change&50M&80M&\shortstack[r]{520M first\\experiment}&\shortstack[r]{520M second\\experiment}\\
\midrule
$16\to8$&$-0.007$&0.104&0.524&0.509\\
$8\to4$&$-0.227$&$-0.136$&$-0.074$&--\\
\bottomrule
\end{tabular}
\end{table}

\subsection{Final-loss differences within independently scheduled runs}\label{app:matched-exposure-controls}

{\color{ReviewPurple}
The following controls were analyzed after the main results and compare recorded training exposure and schedule length.
}

Each repetition count has its own total-token horizon and a cosine schedule defined over that horizon. Final loss therefore changes jointly with the amount of target data consumed and the schedule used to reach that amount. To make this distinction explicit, let

\begin{itemize}
    \item $x$ denote realized target exposure, measured as consumed target tokens divided by the fixed target-pool size;
    \item $R$ denote the target exposure specified for the end of the training configuration, which also determines its total-token and cosine-schedule horizon; and
    \item $\ell_c(N,x;R)$ denote held-out loss on evaluation corpus $c$ (target or generic) for a model of size $N$, observed at exposure $x$ in a run whose final exposure is $R$.
\end{itemize}

For two final exposures $a<b$, let $x_a$ and $x_b$ be their realized final exposures. In the longer run, we also evaluate the saved intermediate model state at exposure $x_a$. It is matched to the final state of the shorter run by random seed, optimization step, total processed tokens, and processed target tokens. The difference between the two final losses can then be written exactly as
\begin{align}
    \ell_c(N,x_b;b)-\ell_c(N,x_a;a)
    &=\underbrace{\ell_c(N,x_a;b)-\ell_c(N,x_a;a)}_{\text{difference at matched exposure}} \nonumber\\
    &\quad+\underbrace{\ell_c(N,x_b;b)-\ell_c(N,x_a;b)}_{\text{difference from additional training}}.
    \label{eq:app-matched-exposure}
\end{align}
The first term compares two runs at the same exposure but with learning-rate schedules defined for different final token counts. The second measures the change when training continues within the longer run from exposure $x_a$ to $x_b$. All differences are ``longer minus shorter,'' so negative values indicate lower loss under the longer configuration.

Table~\ref{tab:app-matched-exposure} reports paired-seed means for Proof-Pile-2 held-out target and generic losses. The target difference at matched exposure becomes small near the high-repetition end, while the difference from additional training changes from negative to positive. Both differences remain negative for held-out generic loss in every displayed comparison.

\begin{table}[H]
    \centering
    \caption{Decomposition at matched target-token exposure on Proof-Pile-2. ``Matched'' compares schedules at the same exposure, ``Additional'' measures further training within the longer run, and ``Final'' is their sum. Negative values favor the longer configuration.}
    \label{tab:app-matched-exposure}
    \normalsize
    \setlength{\tabcolsep}{4pt}
    \begin{tabular}{ccrrrrrr}
        \toprule
        $N$ & $a\to b$ & \multicolumn{3}{c}{Target loss} & \multicolumn{3}{c}{Generic loss} \\
        \cmidrule(lr){3-5}\cmidrule(lr){6-8}
        & & Matched & Additional & Final & Matched & Additional & Final \\
        \midrule
        35M & $6\to12$  & $-0.079$ & $-0.051$ & $-0.130$ & $-0.288$ & $-0.245$ & $-0.533$ \\
        35M & $12\to24$ & $-0.009$ & $0.029$  & $0.020$  & $-0.137$ & $-0.161$ & $-0.298$ \\
        35M & $24\to48$ & $0.002$  & $0.145$  & $0.147$  & $-0.062$ & $-0.106$ & $-0.168$ \\
        50M & $4\to8$   & $-0.139$ & $-0.088$ & $-0.227$ & $-0.339$ & $-0.326$ & $-0.665$ \\
        50M & $8\to16$  & $-0.034$ & $0.027$  & $-0.007$ & $-0.193$ & $-0.202$ & $-0.395$ \\
        50M & $16\to32$ & $0.002$  & $0.188$  & $0.190$  & $-0.087$ & $-0.130$ & $-0.217$ \\
        \bottomrule
    \end{tabular}
\end{table}

The $50$M comparison from $r=8$ to $r=16$ illustrates why final loss alone can hide the transition. At matched exposure, the schedule defined for the longer run improves target loss by 0.034 nat, while additional training to its final token count raises it by 0.027 nat. Their sum still slightly favors $r=16$. At the next interval, the matched-exposure improvement is effectively zero and the target loss added during further training dominates. Meanwhile, held-out generic loss continues to improve.

The Wikipedia-derived target gives the same contrast more weakly. At 50M, increasing the final count from 32 to 48 raises final target loss by 0.033 nat while lowering held-out generic loss by 0.020 nat; all three paired seeds have these signs. The 35M target loss is closer to a plateau and does not show an equally clear increase during further training.

These measurements show that the target-loss increase at high repetition is not accompanied by an increase in held-out generic loss. They do not identify a unique cause. Repetition, target-token fraction, total tokens, and schedule interact in the completed configurations, so the increasing term in Equation~\ref{eq:two-effect-model} should not be labeled as memorization or overfitting without an experiment that isolates those alternatives.

\subsection{Fixed-token and fixed-schedule control}\label{app:fixed-horizon-controls}

The completed configurations above use a different total token count and cosine-schedule horizon for each repetition count. To test whether horizon length alone explains the higher target loss at high repetition, we run a separate Proof-Pile-2 control at 50M parameters. All three configurations use 4,000,317,440 training tokens, the same 4B-token cosine horizon, the same optimizer and target/generic corpora, and paired seeds 1337--1339. The comparison changes the target-token fraction and therefore changes target repetitions and generic-token allocation together.

\begin{table}[H]
    \centering
    \caption{Final losses in the fixed-horizon control. Actual target-token fractions and repetitions differ from the nominal values by less than 1\%.}
    \label{tab:app-fixed-horizon-means}
    \normalsize
    \begin{tabular}{lrrrr}
        \toprule
        Configuration & Target fraction & Repetitions & Target loss & Generic loss \\
        \midrule
        High target fraction & $\approx0.8$ & $\approx32$ & 0.978 & 4.262 \\
        Middle target fraction & $\approx0.4$ & $\approx16$ & 0.748 & 3.904 \\
        Low target fraction & $\approx0.2$ & $\approx8$ & 0.748 & 3.835 \\
        \bottomrule
    \end{tabular}
\end{table}

For the main comparison, define the paired reduction as the target loss at $h\approx0.8,r\approx32$ minus the target loss at $h\approx0.4,r\approx16$. The three reductions are 0.223, 0.244, and 0.222 nat, for a mean of 0.230 nat. All three target-loss reductions are positive, and the corresponding generic-loss reductions are also positive for all three seeds.

A further reduction from $h\approx0.4,r\approx16$ to $h\approx0.2,r\approx8$ does not reliably improve target loss. The three paired target-loss differences, $L_{\mathrm{target}}(16)-L_{\mathrm{target}}(8)$, are $-0.005$, $-0.004$, and $+0.007$ nat, with a negative mean of magnitude $<0.001$ nat. Thus both lower-target configurations improve substantially over the high-target configuration, but their internal ordering is unresolved.

Because total tokens and the cosine horizon are identical, the control rules out horizon length alone as a sufficient explanation for the higher target loss at high repetition. It does not isolate repetition: reducing the target-token fraction also increases the generic-token allocation. The result therefore shows that the combination of high target fraction and high repetition is associated with higher held-out target loss, without identifying memorization, competition between corpus categories, or another unique mechanism.

\section{Candidate Pruning and Prospective Evaluations}\label{app:new-experiments}

{\color{ReviewPurple}
Training configurations and data splits are given in Sections~\ref{app:pubmed-caselaw-settings} and~\ref{app:200m-settings}. This appendix reports the selection rule, completed target-size evaluations, sensitivity to smaller-model inputs, and statistical and compute comparisons.
}

\subsection{Candidate selection by bootstrap and extrapolation}\label{app:pruning-procedure}

{\color{ReviewPurple}
At each smaller model size $N_i$, resample three whole seven-count seed
curves with replacement and average them to obtain $\overline L_i^{(b)}(r)$.
There are $3^3=27$ ordered resamples per size. Combining them independently
across three sizes gives $B=27^3=19{,}683$ combinations, indexed by $b$.
For each combination and count, define
\[
 E_i^{(b)}(r)=\overline L_i^{(b)}(r)
       -\min_{u\in\mathcal R}\overline L_i^{(b)}(u).
\]
Regress these excess losses on $\log N_i$ across the three sizes,
separately for each $(b,r)$. Let the fitted coefficients be
$\widehat\alpha_r^{(b)}$ and $\widehat\beta_r^{(b)}$. The score at target size $N_t$ is
\[
 s_{\mathrm{prune}}(r)=\frac1B\sum_{b=1}^B
 \max\{0,\widehat\alpha_r^{(b)}+\widehat\beta_r^{(b)}\log N_t\}.
\]
Thus clipping precedes averaging. Round scores to $10^{-9}$, retain the
three lowest-scoring counts, and favor smaller $r$ in a tie.
The 520M evaluation uses $N_t=517{,}137{,}408$. Resampling preserves each
seed's complete curve but does not add independent training runs.
}
The retained sets, fixed before any 520M run began, are
$S_{\mathrm{PubMed}}=\{4,8,12\}$ and
$S_{\mathrm{Caselaw}}=\{8,12,16\}$.
The structural law in Section~\ref{sec:structural-regret} was fitted
afterwards and is not the rule used in this prospective experiment.

We retain the three counts with the lowest averaged scores. The baseline
that keeps the three lowest-loss 80M counts gives $\{12,16,24\}$ on PubMed
and $\{16,24,32\}$ on Caselaw. The fixed comparison set is $\{8,12,32\}$;
the swapped-corpus comparison applies each corpus's selected set to the other.

\subsection{Complete 520M results}\label{app:pruning-520m-results}

\begin{table}[H]
\centering\normalsize
\caption{Completed 520M test curves on the additional corpora.
Entries are seven-seed mean $\pm$ standard deviation in nats.
The preselected retained counts are bold.}
\label{tab:new-target-curves}
\begin{tabular}{rcc}
\toprule
$r$ & PubMed & Caselaw\\
\midrule
4  & $\mathbf{3.775}\pm0.151$ & $3.802\pm0.138$\\
8  & $\mathbf{2.339}\pm0.010$ & $\mathbf{2.707}\pm0.062$\\
12 & $\mathbf{2.552}\pm0.013$ & $\mathbf{2.649}\pm0.014$\\
16 & $3.194\pm0.022$ & $\mathbf{3.307}\pm0.042$\\
24 & $4.720\pm0.017$ & $5.141\pm0.023$\\
32 & $5.243\pm0.027$ & $5.775\pm0.017$\\
48 & $5.717\pm0.012$ & $6.263\pm0.014$\\
\bottomrule
\end{tabular}
\end{table}
The selected triples descend and then rise in all seven PubMed seeds and
six of seven Caselaw seeds. The full-grid statistical comparisons and
their interpretation are retained in Section~\ref{app:pruning-statistical}.

{\color{ReviewPurple}
Table~\ref{tab:new-search-benefit} compares the resulting 520M loss with the smaller-model choices and fixed comparison sets.
}

\begin{table}[htb]
\centering\normalsize
\caption{520M loss reduction from using the best count in each
preselected set, relative to the best count in each baseline choice
or set. All values are in nats.}
\label{tab:new-search-benefit}
\begin{tabular}{lrr}
\toprule
Comparison&PubMed&Caselaw\\
\midrule
Reuse the 80M winner&0.854&2.492\\
Keep the three lowest-loss 80M counts&0.213&0.658\\
Use the fixed set $\{8,12,32\}$&0&0\\
Exchange the selected sets between corpora&0&0\\
\bottomrule
\end{tabular}
\end{table}

\subsection{Intermediate-size validation and additional repetition counts}\label{app:intermediate-validation}

The 200M training and evaluation settings are specified in Section~\ref{app:200m-settings}.
The 200M study has two parts: predictions and candidate sets fixed before
target training, followed by additional measurements near the observed
minimum. We retain the original predictions when evaluating the larger
grid. Subsequent fits and detailed curve diagnostics in
Section~\ref{app:intermediate-model-checks} are retrospective analyses of these completed experiments.

\subsubsection{All completed 200M test results}\label{app:200m-complete-results}

\begin{table}[!htbp]
\centering
\caption{All completed 200M target-test results, in nats. Values are
means and sample standard deviations across three seeds, not confidence
intervals. ``Planned'' denotes the additional counts prepared before
target results; ``exploratory'' denotes counts chosen after the main
results. The lowest measured mean in each corpus is bold.}
\label{tab:app200-all-counts}
\normalsize
\begin{tabular}{rll l}
\toprule
$r$ & PubMed mean $\pm$ SD & Caselaw mean $\pm$ SD & Stage \\
\midrule
4  & $2.752\pm0.060$ & $3.229\pm0.062$ & Original \\
6  & $2.474\pm0.002$ & $2.784\pm0.009$ & Planned \\
8  & $2.375\pm0.002$ & $2.610\pm0.009$ & Original \\
9  & $2.348$ (SD $<0.001$) & $2.567\pm0.006$ & Exploratory \\
10 & $2.334$ (SD $<0.001$) & $2.537\pm0.004$ & Planned \\
11 & $\mathbf{2.332}\pm0.003$ & $2.523\pm0.003$ & Exploratory \\
12 & $2.338\pm0.003$ & $\mathbf{2.515}\pm0.002$ & Original \\
14 & $2.367$ (SD $<0.001$) & $2.534\pm0.004$ & Planned \\
16 & $2.438\pm0.003$ & $2.588\pm0.010$ & Original \\
24 & $2.934\pm0.009$ & $3.087\pm0.009$ & Original \\
32 & $3.730\pm0.054$ & $3.872\pm0.028$ & Original \\
48 & $5.234\pm0.018$ & $5.441\pm0.005$ & Original \\
\bottomrule
\end{tabular}
\end{table}

{\color{ReviewPurple}
On $G_7$, PubMed's primary prediction has a paired loss difference
$L(8)-L(12)=0.038\pm0.003$ nats (mean $\pm$ sample SD across three seeds).
The planned refinement gives $L(10)-L(12)=-0.003\pm0.003$ nats.
Each of the six seed curves decreases and then increases on $G_{12}$.
}
The PubMed seedwise minima occur at $11,10,11$; all three Caselaw minima
occur at 12. PubMed's paired difference $L(10)-L(11)$ is
$0.003\pm0.003$ nats, whereas $L(11)-L(12)$ is
$-0.006\pm0.001$ nats. The corresponding Caselaw differences are
$0.014\pm0.001$ and $0.008\pm0.003$ nats. For these paired differences, uncertainty is the sample SD of the three differences.
These observations locate a relatively flat PubMed minimum near
$r=10$--12; they do not determine the optimum at unmeasured counts.
In particular, Caselaw $r=13$ was not evaluated.

\subsubsection{All predictions and retained sets fixed before target training}\label{app:200m-frozen-predictions}

Table~\ref{tab:app200-frozen-functions} complements the primary comparison
in Table~\ref{tab:intermediate-predictions}. The secondary fits are the
additive model with free $q$ and the saturated model of
Equation~\ref{eq:saturated-two-effect}, both fitted only to the same
35M/50M/80M sources. The saturated fit uses exponent upper bounds of 4.
The auxiliary analysis uses only 35M and 50M. None of these alternatives
replaces the primary prediction after seeing the target results.

\begin{table}[!htbp]
\centering
\caption{All frozen point predictions. \rev{$r_*(N)$} is the continuous
function minimum; $r_7$ is the count selected on the original grid.
$\Delta_7$ and $\Delta_{12}$ are the measured excess losses of this
\emph{same} $r_7$ relative to the lowest mean on $G_7$ and $G_{12}$.
Local RMSE compares predicted and measured $L(r)-L(12)$ on
$\{8,9,10,11,12,14,16\}$.}
\label{tab:app200-frozen-functions}
\normalsize
\setlength{\tabcolsep}{4pt}
\begin{tabular}{llrrrrr}
\toprule
Corpus & Frozen fit & \rev{$r_*(N)$} & $r_7$ & $\Delta_7$ & $\Delta_{12}$ & Local RMSE \\
\midrule
PubMed & Primary, $q=1$ & 9.659 & 8 & 0.038 & 0.044 & 0.021 \\
 & Free $q$ & 7.730 & 8 & 0.038 & 0.044 & 0.075 \\
 & Saturated & 14.827 & 16 & 0.100 & 0.106 & 0.048 \\
 & 35M/50M, $q=1$ & 10.534 & 12 & 0 & 0.006 & 0.015 \\
\midrule
Caselaw & Primary, $q=1$ & 11.431 & 12 & 0 & 0 & 0.019 \\
 & Free $q$ & 10.112 & 12 & 0 & 0 & 0.044 \\
 & Saturated & 16.313 & 16 & 0.073 & 0.073 & 0.046 \\
 & 35M/50M, $q=1$ & 12.107 & 12 & 0 & 0 & 0.016 \\
\bottomrule
\end{tabular}
\end{table}

The frozen retained sets are $\{8,12,16\}$ for PubMed and
$\{12,16,24\}$ for Caselaw; the auxiliary two-source application gives
the same sets. Both retain the lowest-loss count on $G_7$, for each seed
and for the mean. PubMed's retained-set excess loss changes from 0
on $G_7$ to 0.003 after adding $\{6,10,14\}$ and to 0.006 on $G_{12}$.
Caselaw's remains 0. The fixed comparison set $\{8,12,32\}$ has
the same retained-set excess losses in all three stages. This is evidence
that the rule locates useful candidates, without an observed loss
advantage over that fixed set. There is no general theoretical guarantee
that pruning reduces total computation.

On the common test split and $G_7$, the preferred counts at
35M/50M/80M/200M/520M are $48/32/16/12/8$ for PubMed and
$48/32/24/12/12$ for Caselaw. At 35M the minimum is the largest evaluated
count, so an interior minimum is not observed. Reusing the count preferred
by the original 80M selection curves gives 200M excess losses of
0.106 and 0.571 nats on $G_{12}$, compared with 0.044 and 0 for
the primary frozen point predictions.

\subsection{Candidate pruning with fewer source seeds or sizes}

\label{app:pruning-source-sensitivity}

We reproduce the \rev{candidate-pruning rule from Section~\ref{sec:candidate-pruning}},
including all seven original scores and the frozen 520M sets. The rule
resamples source curves, measures each count's loss relative to the
source minimum, extrapolates these differences against log model size,
and retains three candidates. This rule is distinct from the parametric
models in Section~\ref{app:additional-comparisons}.

At each source size, we keep one, two, or all three existing seeds and
enumerate every subset of the chosen size. For a 520M target we use
35/50/80, 35/50, 35/80, and 50/80 as sources. A three-size source yields
$27+27+1=55$ seed combinations, and each two-size source yields
$9+9+1=19$. Thus each corpus has 112 variants for 520M. Using 35/50 to
predict 80M adds 19 per corpus, giving 262 variants in total. No new
source models are trained in this analysis.

\begin{table}[htbp]
 \centering
 \caption{Retained sets using all three available seeds at each listed
 source size. Every row retains the lowest target mean loss in both
 corpora. The full sensitivity analysis also includes seed subsets.}
 \label{tab:app-source-sets}
 \normalsize
 \begin{tabular}{llll}
  \toprule
  Target (M) & Sources (M) & PubMed & Caselaw \\
  \midrule
  520 & 35, 50, 80 & $\{4,8,12\}$ & $\{8,12,16\}$ \\
  520 & 35, 50 & $\{8,12,16\}$ & $\{8,12,16\}$ \\
  520 & 35, 80 & $\{4,8,12\}$ & $\{8,12,16\}$ \\
  520 & 50, 80 & $\{4,8,12\}$ & $\{8,12,16\}$ \\
  80 & 35, 50 & $\{16,24,32\}$ & $\{16,24,32\}$ \\
  \bottomrule
 \end{tabular}
\end{table}

All 112 variants per corpus at 520M retain both $r=8$ and $r=12$,
including the minimum of the target means and the best count of every
individual target seed (seven seeds per corpus). All 19 variants per
corpus at 80M retain its minimum mean and the best count of every
target seed (three per corpus). The outer candidate can change:
Caselaw's three-size subsets select either $\{4,8,12\}$ or
$\{8,12,16\}$, while some PubMed subsets for 80M select
$\{12,16,24\}$. The low-loss candidates remain present.

\begin{table}[htbp]
 \centering
 \caption{Lowest target mean loss available within each set, in nats.
 The fixed comparison set is $\{8,12,32\}$. The 80M rows retain the
 original selection-split evaluation; the 520M rows use test loss.
 Three source seeds per size are used.}
 \label{tab:app-fixed-set-loss}
 \normalsize
 \begin{tabular}{llrrr}
  \toprule
  Corpus & Target (M) & Pruned set & Fixed set & Fixed minus pruned \\
  \midrule
  PubMed & 80 & 2.311 & 2.354 & 0.043 \\
  Caselaw & 80 & 2.541 & 2.603 & 0.062 \\
  PubMed & 520 & 2.339 & 2.339 & 0 \\
  Caselaw & 520 & 2.649 & 2.649 & 0 \\
  \bottomrule
 \end{tabular}
\end{table}

At 520M the fixed and pruned sets reach the same measured loss; at 80M
the retained sets reach lower loss than this fixed comparison. These
are set-quality comparisons, based on the best mean within each set.
Candidate pruning has no general theoretical guarantee of lower total
training cost.

The source-subset analyses assess robustness to removing existing seeds
and scales. They do not constitute independent repetitions with new
seeds, and the analysis was specified after the target results were
available. Target losses are not used to generate the sets. The
Wikipedia-derived and Proof-Pile-2 source grids differ across sizes and
do not support the same three-candidate, count-by-count pruning test:
the former has only three counts common to all source sizes and the
latter only one common to 35M and 50M. We therefore keep their
parametric-model analyses separate from this pruning sensitivity check.

\subsection{Statistical comparisons of retained sets}\label{app:pruning-statistical}

The model regret and the measured loss improvement answer different
questions. Regret relative to an unmeasured continuous minimum depends
on the fitted structure. The loss difference between two completed
configurations does not. Both are useful: the former gives a conjectural
cost for coarse discretization, while the latter measures the benefit
of reaching the observed low-loss region.

We also report the prespecified finite-grid comparison. Let $\mu(r)$ be
the expected loss over training seeds on the fixed 520M test split,
$\mathcal R=\{4,8,12,16,24,32,48\}$, and $S\subset\mathcal R$ the retained set.
For each ordered pair, $U(a,q)$ is the upper endpoint for $\mu(a)-\mu(q)$
from seven paired seeds. The analysis uses 84 directed comparisons across
the two corpora, a Bonferroni-adjusted Student-$t$ critical value of 5.763
with six degrees of freedom, and 95\% simultaneous coverage under the
independent-seed, normally distributed paired-difference model.
The prespecified sufficient upper bound is
\[
 B_{\mathrm{original}}(S)=\min_{a\in S}\max_{q\in\mathcal R}U(a,q),
 \qquad U(a,a)=0.
\]
This bound is 0 on PubMed
and 0.089 nat on Caselaw, so its joint 0.02-nat criterion is not met.
The Caselaw bound is driven by uncertainty between the two retained
counts 8 and 12, rather than by a lower-loss discarded count.
An exact set-regret identity,
\[
 \rho(S)=\max\left\{0,\max_{q\in\mathcal R\setminus S}\min_{a\in S}
                                  [\mu(a)-\mu(q)]\right\},
\]
gives the supplementary bound
\[
 \rho(S)\le\max\left\{0,\max_{q\in\mathcal R\setminus S}\min_{a\in S}U(a,q)\right\}.
\]
Using the same simultaneous intervals yields zero for both preselected sets. This set-level
reanalysis is post hoc. Fixed and swapped-corpus sets also retain the
lowest-loss measured counts, so these experiments do not establish an
additional benefit unique to selecting on the same corpus.

\subsection{Compute and recorded running time}\label{app:pruning-cost}

Keeping three of seven counts removes 57.1\% of candidate configurations.
Because the removed counts have longer horizons, evaluating only the
preselected target sets would reduce the 520M-stage $6ND$ cost by 83.3\%
on PubMed and 75.0\% on Caselaw. Including the full smaller-model scans gives
a combined reduction of 65.6\% relative to directly running both complete
520M sweeps. This is a cost calculation for deploying the preselected rule;
the validation experiment actually completed all 520M counts.
The recorded single-GPU completed-run durations sum to 1074.9 hours.
Using those durations instead of the $6ND$ proxy gives an estimated
59.9\% reduction relative to the 520M-only full sweeps, including
the smaller-model runs. Reservation idle time and other jobs are not included.

The recorded times are about 173.6 single-GPU hours for all smaller-model
runs and 901.3 hours for both complete 520M grids. The smaller-model runs
plus the retained 520M counts would take about 361.4 hours at these measured
durations. All cost comparisons use the same recorded seed allocation.

{\color{ReviewPurple}
The pruned sets together with all smaller-model scans have a combined
$6ND$ estimate 4.7\% below training only the fixed set $\{8,12,32\}$
at 520M, with the same lowest measured loss. The estimated cost is lower
on PubMed and higher on Caselaw. Using the recorded durations, the fixed
set takes 325.2 GPU hours versus 361.4 for the smaller-model scans plus
pruned sets, an 11.1\% increase. The small $6ND$ advantage therefore does
not imply a running-time advantage. These comparisons concern the
original 520M study; they exclude the later 200M study and finer-grid
runs. They estimate applying the rule, rather than savings realized by
the full validation sweeps.
}

\section{Mathematical Details}\label{app:K}

\begin{revision}
This appendix reports the observed reversal decomposition from
Section~\ref{sec:rankings} and proves the statements used in
Sections~\ref{sec:rankings} and~\ref{sec:scale-model}. The losses are
real-valued functions on a finite set of evaluated repetition counts,
so the results do not identify a continuous optimum.
\begin{table}[htb]
\color{Revision}
\centering\normalsize
\caption{Decomposition of the observed reversals, in nats. Both counts
are measured at each size. Entries are computed before rounding.}
\label{tab:rank-decomposition}
\begin{tabular}{lccrrr}
\toprule
Corpus & Sizes & $b\to a$ & $\delta(b)-\delta(a)$ & $m_s$ & $G_{b\to a}$\\
\midrule
Wikipedia & 35M$\to$80M & $36\to24$ & 0.203 & 0.064 & 0.139\\
Proof-Pile-2 & 50M$\to$80M & $16\to8$ & 0.110 & 0.007 & 0.104\\
\bottomrule
\end{tabular}
\end{table}
\end{revision}

Let $\mathcal R$ be a finite candidate set with a deterministic tie rule. Write $L_s,L_t:\mathcal R\to\mathbb R$ for the smaller- and larger-model loss curves, and let $r_s$ and $r_t$ minimize them. Define
\[
g_s(r)=L_s(r)-L_s(r_s)\geq0,
\qquad
\delta(r)=L_t(r)-L_s(r).
\]

\paragraph{Proposition~\ref{app:K}.1 (exact decomposition).}
For every $r\in\mathcal R$,
\begin{equation}
L_t(r_s)-L_t(r)=\delta(r_s)-\delta(r)-g_s(r).
\label{eq:app-exact-decomposition}
\end{equation}
Thus $r$ beats the smaller-model choice at the larger scale exactly when $\delta(r_s)-\delta(r)>g_s(r)$.

\emph{Proof.} Substitute $L_t=L_s+\delta$ and collect the smaller-model terms. $\square$

\paragraph{Proposition~\ref{app:K}.2 (nearly common cross-scale changes).}
Let $\operatorname{osc}(\delta)=\max_r\delta(r)-\min_r\delta(r)$. Then
\[
L_t(r_s)-L_t(r_t)\leq\operatorname{osc}(\delta).
\]
Equivalently, if all cross-scale changes lie within $\varepsilon$ of one constant, the loss increase from reuse is at most $2\varepsilon$.

\emph{Proof.} Apply Equation~\ref{eq:app-exact-decomposition} with $r=r_t$, use $g_s(r_t)\geq0$, and note that the best constant is the midpoint of the largest and smallest $\delta(r)$. $\square$

\paragraph{Proposition~\ref{app:K}.3 (smaller-model fit error is not enough).}
Suppose a fitted curve $\widehat L_s$ obeys $\lVert\widehat L_s-L_s\rVert_\infty\leq\zeta$, and let $\widehat r$ minimize $\widehat L_s$. Then
\[
L_s(\widehat r)-L_s(r_s)\leq2\zeta,
\qquad
L_t(\widehat r)-L_t(r_t)\leq2\zeta+\operatorname{osc}(\delta).
\]

\emph{Proof.} Optimality under the fitted curve gives
$L_s(\widehat r)\leq\widehat L_s(\widehat r)+\zeta\leq\widehat L_s(r_s)+\zeta\leq L_s(r_s)+2\zeta$.
Adding the largest possible difference in $\delta$ gives the second inequality. $\square$

{\color{ReviewPurple}
For an equally weighted root-mean-square error $e$ on the same finite set, $\lVert\widehat L_s-L_s\rVert_\infty\leq\sqrt{|\mathcal R|}\,e$. Thus RMSE bounds the uniform error after accounting for the number of candidates. Even a perfect smaller-model fit leaves $\operatorname{osc}(\delta)$ unconstrained, so fit quality at one scale alone cannot guarantee transfer.
}

\paragraph{Proposition~\ref{app:K}.4 (pairwise ordering under the two-effect model).}
Suppose that, for a fixed target corpus and a local range of model sizes and counts,
\[
L_N(r)=A(N)+N^{\eta}U(r)+N^{\eta+q}V(r),
\qquad U'(r)<0,\quad V'(r)>0,\quad q>0.
\]
For any $a<b$, the ordering between $a$ and $b$ can reverse at most once as $N$ increases. If $L_{N_0}(a)\leq L_{N_0}(b)$, then $L_N(a)<L_N(b)$ for every $N>N_0$ in the range, apart from equality already allowed at $N_0$.

\emph{Proof.} Let $C=U(a)-U(b)>0$ and $D=V(b)-V(a)>0$. Then
\[
N^{-\eta}[L_N(b)-L_N(a)]=DN^q-C,
\]
which is strictly increasing for $q>0$. Its sign can change at most once and cannot return after becoming nonnegative. $\square$

The increasing quantity is the normalized difference $N^{-\eta}[L_N(b)-L_N(a)]$. The proof does not require the unnormalized difference itself to grow.

\paragraph{Proposition~\ref{app:K}.5 (selection on one fixed candidate set).}
If Proposition~\ref{app:K}.4 holds for every pair in a fixed finite set $\mathcal R$ and ties are broken toward the smaller count, then the selected count \rev{$r_N=\arg\min_{r\in\mathcal R}L_N(r)$} is weakly decreasing in $N$.

\emph{Proof.} Let $N_2>N_1$ and \rev{$r_1=r_{N_1}$}. Every $b>r_1$ is no better than $r_1$ at $N_1$, and Proposition~\ref{app:K}.4 prevents it from becoming better at $N_2$. The smallest minimizer at $N_2$ can therefore be no larger than $r_1$. $\square$

The fixed-set condition matters. Proof-Pile-2 uses different candidate sets at 35M, 50M, and 80M, so the observed sequence $12\to16\to8$ does not contradict this result. Different finite candidate sets can map a moving continuous minimum to a nonmonotone sequence; this is a possible explanation, not an identified cause.

\paragraph{Proposition~\ref{app:K}.6 (continuous minimum of a local power-law form).}
Consider the specialization
\begin{equation}
L_N(r)=A(N)+B_{\mathrm{dec}}N^{\eta}r^{-\alpha}+P_{\mathrm{inc}}N^{\eta+q}r^{\nu},
\qquad B_{\mathrm{dec}},P_{\mathrm{inc}},\alpha,\nu,q>0.
\label{eq:app-local-power-law}
\end{equation}
Writing $z=\log r$, its second derivative with respect to $z$ is
\[
\alpha^2B_{\mathrm{dec}}N^{\eta}e^{-\alpha z}
+\nu^2P_{\mathrm{inc}}N^{\eta+q}e^{\nu z}>0.
\]
The loss is strictly convex in $\log r$ and has the unique continuous minimum
\begin{equation}
r^*(N)=
\left(\frac{\alpha B_{\mathrm{dec}}}{\nu P_{\mathrm{inc}}}\right)^{1/(\alpha+\nu)}
N^{-q/(\alpha+\nu)}.
\label{eq:app-continuous-minimum}
\end{equation}
This formula explains how a local continuous minimum can move downward when $q>0$. It does not predict the globally lowest-loss 520M count: extrapolating the fitted 35M/50M function favors $r=4$, whereas the measured 520M ordering is $L_{520\mathrm{M}}(8)<L_{520\mathrm{M}}(4)<L_{520\mathrm{M}}(16)$.

{\color{ReviewPurple}
\subsection{Local approximation in log model size}\label{app:local-pruning}
Consider a common finite candidate set $\mathcal R$. Suppose $r_0$ remains
a minimizer near a reference size $N_0$. A unique minimizer at $N_0$
ensures such a neighborhood under Equation~\ref{eq:two-effect-model},
since there are finitely many continuous pairwise loss differences.
Writing $x=\log(N/N_0)$, define
\[
 C_r=N_0^\eta[U(r)-U(r_0)],\qquad
 D_r=N_0^{\eta+q}[V(r)-V(r_0)].
\]
Subtracting $L_N(r_0)$ removes $A(N)$ and gives
\[
 E_N(r)=C_re^{\eta x}+D_re^{(\eta+q)x}
       =c_r+d_rx+O(x^2),
 \quad c_r=C_r+D_r,\quad d_r=\eta C_r+(\eta+q)D_r.
\]
Hence the first-order term has the form $\alpha_r+\beta_r\log N$, with
$\alpha_r=c_r-d_r\log N_0$ and $\beta_r=d_r$.
For $|x|\leq d_x$ within this neighborhood, Taylor's formula bounds the remainder by
\[
 |E_N(r)-c_r-d_rx|
 \leq \frac{x^2}{2}\left[
 \eta^2|C_r|e^{|\eta|d_x}
 +(\eta+q)^2|D_r|e^{|\eta+q|d_x}\right].
\]
The coefficients and neighborhood width are unknown; this is not a
numerical error bound for our target-scale predictions. The expansion
also exists for either sign of $q$. Positivity of $q$ determines the
ranking-change direction, not the existence of a local linear approximation.

When the minimizing candidate changes,
$E_N(r)=\max_{u\in\mathcal R}[L_N(r)-L_N(u)]$ can have a corner.
Linearizing the pairwise differences then gives a maximum of affine
functions, generally not a single line across the switch. The observed
smaller-model minima change from 48 to 32 to 16 on PubMed and from 48 to
32 to 24 on Caselaw, so one fixed-minimizer neighborhood does not cover
the full source range. The local expansion explains the regression form;
it does not derive bootstrap resampling, clipping, or the number of
retained candidates, nor guarantee distant extrapolation. Other smooth
loss models also admit local linear approximations.

}

\section{Empirical Models and Numerical Checks}\label{app:empirical-models}

\begin{revision}
Fits allowing $q$ to cross zero give positive estimates on all four
corpora. Using only 35M--80M with $\eta=0$, PubMed and Caselaw give
$q=1.309$ and $1.175$, with seed-bootstrap intervals above zero.
Appendix~\ref{app:q-sign-check} reports the fitted forms and uncertainty.

Matched-exposure and fixed-horizon controls suggest that schedule length
alone does not explain the high-repetition loss increase. They do not
separate repetition from mixture composition and generic exposure
(Appendices~\ref{app:matched-exposure-controls}
and~\ref{app:fixed-horizon-controls}).
\end{revision}

\subsection{Simple fit to the evaluated 80M choice}\label{app:posthoc-controls}

{\color{ReviewPurple}
This retrospective fit uses completed smaller-model curves to check the evaluated 80M choice.
}

\begin{revision}
For each target corpus, we fit Equation~\ref{eq:app-posthoc-fit}
using the completed 35M and 50M loss curves and evaluate it on the
existing 80M candidate set. Each source size has a separate intercept;
the two repetition coefficients are shared and the powers are fixed.
\end{revision}

Table~\ref{tab:app-posthoc-fit} gives the results using both the nominal model names and the realized parameter counts. Both target studies use the same model configurations, with realized counts of 34,453,888, 48,826,368, and 80,710,400 parameters at the three nominal scales. Figures and prose use the rounded nominal names; calculations labeled ``realized'' use these exact counts. Changing the convention moves each continuous minimum by about 2\% and does not change either evaluated choice.

\begin{table}[H]
    \centering
    \caption{Post hoc fit to the completed 35M and 50M curves. The arrow in the continuous-minimum column shows the result using nominal parameter counts followed by the result using realized parameter counts.}
    \label{tab:app-posthoc-fit}
    \normalsize
    \setlength{\tabcolsep}{3pt}
    \begin{tabularx}{\textwidth}{@{}>{\raggedright\arraybackslash}p{0.19\textwidth}>{\raggedright\arraybackslash}p{0.205\textwidth}>{\raggedright\arraybackslash}p{0.205\textwidth}>{\raggedleft\arraybackslash}p{0.19\textwidth}>{\raggedleft\arraybackslash}X@{}}
        \toprule
        Target corpus & Nominal $(B_{\mathrm{dec}},P_{\mathrm{inc}})$ & Realized $(B_{\mathrm{dec}},P_{\mathrm{inc}})$ & 80M continuous minimum & Evaluated choice \\
        \midrule
        Wikipedia-derived & $(4.431,\,0.011)$ & $(4.431,\,0.011)$ & $25.82\to25.32$ & $24\to24$ \\
        Proof-Pile-2 & $(1.690,\,0.018)$ & $(1.683,\,0.017)$ & $9.67\to9.48$ & $8\to8$ \\
        \bottomrule
    \end{tabularx}
\end{table}

We next perform an exhaustive paired bootstrap over the smaller-model seeds. Within each model size, seed identifiers are sampled with replacement, and the same sampled identifiers are applied to every repetition count before refitting Equation~\ref{eq:app-posthoc-fit}; this preserves the within-seed loss curve across counts. The Proof-Pile-2 curves contain four complete per-seed curves at 35M and three at 50M, giving $4^4\times3^3=6912$ ordered bootstrap resamples. {\color{ReviewPurple}
Using the realized parameter counts, every fit has positive $B_{\mathrm{dec}}$ and $P_{\mathrm{inc}}$ and selects $r=8$ in the evaluated 80M candidate set. The continuous minima range from 9.304 to 9.611 over all 6912 resamples; this is the full range, not a 95\% interval.
}

The Wikipedia-derived curves have the same nominal seed allocation, except that seed 1339 is missing at 35M, $r=24$. In each bootstrap resample, the mean for a repetition count uses the sampled observations available at that count. The only unavailable 35M resample is the one that selects seed 1339 four times, because it contains no observation at $r=24$. Pairing it with the 27 ordered 50M resamples accounts for all 27 unavailable fits. The other 6885 fits have positive coefficients and all select $r=24$. No value is imputed for the missing run.

\begin{table}[H]
    \centering
    \caption{Stability of the fitted 80M choices under the paired bootstrap of completed smaller-model curves.}
    \label{tab:app-posthoc-bootstrap}
    \normalsize
    \setlength{\tabcolsep}{3pt}
    \begin{tabularx}{\textwidth}{@{}>{\raggedright\arraybackslash}p{0.23\textwidth}>{\raggedleft\arraybackslash}p{0.16\textwidth}>{\raggedleft\arraybackslash}p{0.15\textwidth}>{\raggedleft\arraybackslash}p{0.15\textwidth}>{\raggedright\arraybackslash}X@{}}
        \toprule
        Target corpus & Ordered bootstrap resamples & Available fits & Unavailable fits & Same 80M choice \\
        \midrule
        Wikipedia-derived & 6912 & 6885 & 27 & 6885/6885 select $r=24$ \\
        Proof-Pile-2 & 6912 & 6912 & 0 & 6912/6912 select $r=8$ \\
        \bottomrule
    \end{tabularx}
\end{table}

For Proof-Pile-2, a separate bootstrap of the three observed 80M paired differences gives 27 ordered resamples. All 27 have positive mean $L_{80\mathrm{M}}(16)-L_{80\mathrm{M}}(8)$. This check concerns the measured 80M comparison and is distinct from refitting the 35M and 50M curves.

\subsection{Structural fits, compute envelopes, and repetition regret}\label{app:structural-regret}

\subsubsection{Why the extension preserves the original argument}

In Equation~\ref{eq:saturated-two-effect}, all shape parameters except $\eta$ are positive; $A_N$ is a separate loss offset at each model size.

More generally, suppose $L_N(r)=A_N+H_N(a_N[U(r)+w(N)V(r)])$,
where $a_N>0$, $H_N'>0$, $U'<0$, and $V'>0$.
For $a<b$, put $C=U(a)-U(b)>0$ and $D=V(b)-V(a)>0$.
The sign of $L_N(b)-L_N(a)$ is the sign of $w(N)D-C$.
An increasing $w$ permits at most one reversal of this sign. Thus an
increasing transformation need not preserve the additive loss difference,
but it does preserve the pairwise ordering used in the original argument.

For a valley, one further condition is needed. Define
$m(r)=-U'(r)/V'(r)$. If $m$ is strictly decreasing and $w(N)$ lies in
its range, the sign of the loss derivative changes once, from negative
to positive, at $m(r_*)=w(N)$. The powers in
Equation~\ref{eq:saturated-two-effect} satisfy this condition:
$m(r)$ is proportional to $r^{-(\alpha+\nu)}$.
The loss need not be globally convex after the increasing saturation.

\subsubsection{Exact regret and local discretization cost}

Let $u=\log(r/r_*)$, $F_*=F_N(r_*)$, and define
\begin{equation}
 Q(u)=\frac{\nu e^{-\alpha u}+\alpha e^{\nu u}}{\alpha+\nu}-1,
 \qquad
 s_N=\frac{K^2F_*}{(K+F_*)^2},\quad
 \lambda_N=\frac{F_*}{K+F_*}.
 \label{eq:regret-shape}
\end{equation}
The first-order balance at $r_*$ gives $F_N(r)=F_*[1+Q(u)]$.
Consequently the model regret is exactly
\begin{equation}
 \widehat R_N(r):=\widehat L_N(r)-\widehat L_N(r_*)
 =\frac{s_N Q\!\left(\log(r/r_*)\right)}
        {1+\lambda_N Q\!\left(\log(r/r_*)\right)}.
 \label{eq:structural-regret}
\end{equation}
Weighted arithmetic--geometric mean implies $Q(u)\ge0$, with equality
only at $u=0$. 

Write $x_*=r_*/8$, $T=n^\eta Bx_*^{-\alpha}$.
At the optimum, $n^\eta Pw(n)x_*^\nu=(\alpha/\nu)T$.
Hence $F_*=T(1+\alpha/\nu)$ and
\[
 \frac{F_N(r)}{F_*}
 =\frac{\nu e^{-\alpha u}+\alpha e^{\nu u}}{\alpha+\nu}
 =1+Q(u),\qquad u=\log(r/r_*).
\]
Subtracting the two saturated losses gives
\[
 \frac{KF_N(r)}{K+F_N(r)}-\frac{KF_*}{K+F_*}
 =\frac{K^2(F_N(r)-F_*)}{(K+F_N(r))(K+F_*)},
\]
which is Equation~\ref{eq:structural-regret}.
The offset disappears exactly. Without loss saturation, the analogous
regret is $F_*Q(u)$, so the original additive model is covered as a limit.

Since $Q(0)=Q'(0)=0$ and $Q''(0)=\alpha\nu$, Taylor expansion
gives Equation~\ref{eq:local-regret}, with $k_N=\alpha\nu s_N$.
For $|u|\le d$,
\[
 Q''(u)=\frac{\alpha\nu}{\alpha+\nu}
                 \bigl[\alpha e^{-\alpha u}+\nu e^{\nu u}\bigr]
 \le\alpha\nu e^{m d},\qquad m=\max(\alpha,\nu).
\]
Taylor's integral formula then gives
$Q(u)\le\alpha\nu e^{m d}u^2/2$.
The denominator in Equation~\ref{eq:structural-regret} is at least one,
so $\widehat R_N(r)\le k_Ne^{m d}u^2/2$.
If the interval brackets $r_*$ and adjacent log-counts are separated
by at most $h_r$, a nearest count has $|u|\le h_r/2$, giving the
stated $k_Ne^{m h_r/2}h_r^2/8$ bound. The local approximation for an
additive grid with spacing $\Delta r$ is
$k_N(\Delta r)^2/(8r_*^2)$; it is an approximation because the
coordinate transformation and curvature vary across the cell.

\paragraph{Cross-scale reuse.}
If one reuses the continuous model optimum from $N_s$ at $N_t>N_s$,
its log-error is
\[
 u_{s\to t}=\log\frac{r_*(N_s)}{r_*(N_t)}
 =\frac{\log w(N_t)-\log w(N_s)}{\alpha+\nu}.
\]
Substituting this into Equation~\ref{eq:structural-regret} gives the
exact model cost of reuse. For a small scale change, this cost is
approximately $k_{N_t}u_{s\to t}^2/2$.
Moving toward the fitted minimum reduces the model regret; a retained
bracket then makes a finer grid useful through the quadratic local cost.
This analysis motivates refinement but does not establish convergence
of an arbitrary noisy gradient algorithm. The rule evaluated here uses
bootstrap resampling of smaller-model curves and evaluates a preselected set at the target size.

\subsubsection{Fitting and checks}\label{app:structural-fitting-checks}

The inputs are 97 mean losses, one per corpus, model size, and repetition configuration: 21 for Wikipedia-derived,
20 for Proof-Pile-2 including its primary 520M comparison, and 28 each
for PubMed and Caselaw. The Proof-Pile-2 replication with new seeds remains
a separate comparison. Each corpus has its own shape parameters; no
shared universal exponent is imposed. At each size we fit a separate
additive offset, and minimize equally weighted squared errors across
these means. This estimates a shape law under the recorded training recipe,
not an absolute-loss law across evaluation splits or a fully tuned
compute frontier.

The central fit constrains $\alpha,\nu,q\le4$. A sensitivity fit
raises these limits to 8, using the same functional family and additional
initializations. Other bounds are $\alpha,\nu\ge0.1$,
$q\ge0.01$, $-2\le\eta\le2$, $0.1\le K\le100$,
$0.2\le n_s\le40$, and $-12\le\log B,\log P\le8$.
Some parameters reach a boundary, so the exponents are not treated as
identified constants. We also withhold each interior count in
$\{8,12,16,24,32\}$ at all four new-corpus sizes, refit using only
the remaining counts, and evaluate its omitted mean loss. These are
retrospective interpolation diagnostics; they do not give the fitted
law the prospective status of the preselected candidate rule.

\begin{table}[t]
\centering\normalsize
\caption{Structural fit error in nats. Held-out columns withhold each of
five interior repetition counts at all four new-corpus scales.
The wider exponent limits improve reconstruction slightly but do not
improve this interpolation diagnostic.}
\label{tab:structural-fit-error}
\begin{tabular}{lrrrr}
\toprule
 & \multicolumn{2}{c}{Fit RMSE} & \multicolumn{2}{c}{Held-out RMSE}\\
Corpus & Cap 4 & Cap 8 & Cap 4 & Cap 8\\
\midrule
Wikipedia-derived & 0.020 & 0.020 & --- & ---\\
Proof-Pile-2 & 0.015 & 0.015 & --- & ---\\
PubMed & 0.089 & 0.088 & 0.107 & 0.113\\
Caselaw & 0.076 & 0.071 & 0.093 & 0.095\\
\bottomrule
\end{tabular}
\end{table}

\begin{table}[t]
\centering\normalsize
\caption{Parameters for evaluating the central 520M regret expression
directly. They summarize the fitted shape and do not use the size offset.}
\label{tab:regret-shape-parameters}
\begin{tabular}{lrrrrr}
\toprule
Corpus & $r_*$ & $\alpha$ & $\nu$ & $s_N$ & $\lambda_N$\\
\midrule
Proof-Pile-2 & 6.026 & 1.938 & 2.383 & 0.275 & 0.233\\
PubMed & 9.490 & 2.279 & 3.734 & 0.602 & 0.175\\
Caselaw & 9.984 & 1.536 & 4.000 & 0.779 & 0.207\\
\bottomrule
\end{tabular}
\end{table}

\begin{table}[t]
\centering\normalsize
\caption{520M quantities derived from the fitted two-effect law.
Regret is relative to the unmeasured continuous model minimum, in nats.
Ranges span the two exponent-bound choices, not confidence intervals.}
\label{tab:continuous-regret}
\begin{tabularx}{\textwidth}{@{}l*{4}{>{\raggedleft\arraybackslash}X}@{}}
\toprule
Corpus & Fitted $r_*$ & $k_N$ (cap 4) & Retained-set regret & Unit-grid regret\\
\midrule
PubMed & 9.46--9.49 & 5.124 & 0.066--0.069 & 0.006--0.007\\
Caselaw & 9.98--10.32 & 4.784 & 0.060--0.094 & $<0.003$\\
\bottomrule
\end{tabularx}
\end{table}

The unit-grid calculation considers integer $r$ from 4 through 48 and
selects 9 on PubMed and 10 on Caselaw. These counts have not been trained.
The central model's 0.02-nat regret sublevel intervals are approximately
$[8.67,10.35]$ and $[9.08,10.90]$, respectively. They are useful refinement
hypotheses, not experimentally certified tolerances.
In particular, the whole-curve fit errors are comparable to the estimated
retained-set regrets. On PubMed the fitted minimum absolute loss is even
above the lowest observed endpoint mean, because a global least-squares
fit need not lie below every measurement. The regret estimates therefore
describe the fitted shape and cannot be substituted for observed or
certified excess loss. Their more stable conclusion is the approximate
valley location and the possibility of a remaining discretization cost.

For the Proof-Pile-2 smaller models, local quadratic interpolation
on repetition and log repetition places the minima near 15--17, 12--13,
and 8--9 at 35M, 50M, and 80M, respectively. This supports a possible
downward movement obscured by the differing grids.

For the first Proof-Pile-2 520M experiment, the structural fit gives
$r_*\approx6.03$ and $k_N\approx1.27$. This places its valley in the
span of the measured counts 4, 8, and 16. On Wikipedia-derived at 80M,
the structural minimum is approximately 18.97. For comparison, local
quadratic interpolation on $r$ and $\log r$ gives 6.16--7.32 on the first
520M Proof-Pile-2 curve and 20.01--20.75 on Wikipedia-derived at 80M.
Those coordinate-dependent estimates are supplementary checks, not the
functions used to construct the structural envelope.

Direct pure-power extrapolation of small-model interpolated valley
locations to 520M remains too aggressive: the estimates are approximately
2.2--2.3 on Proof-Pile-2, 3.8--4.5 on PubMed, and 3.0--3.2 on Caselaw.
The earlier structural fit restricted to small scales also extrapolates
poorly on the two new corpora. Thus the retrospective structural analysis
supports a curved valley trajectory, while the prospective evidence
belongs to the preselected candidate rule, not to a newly validated scaling law.

\begin{figure}[H]
 \centering
 \includegraphics[width=\textwidth]{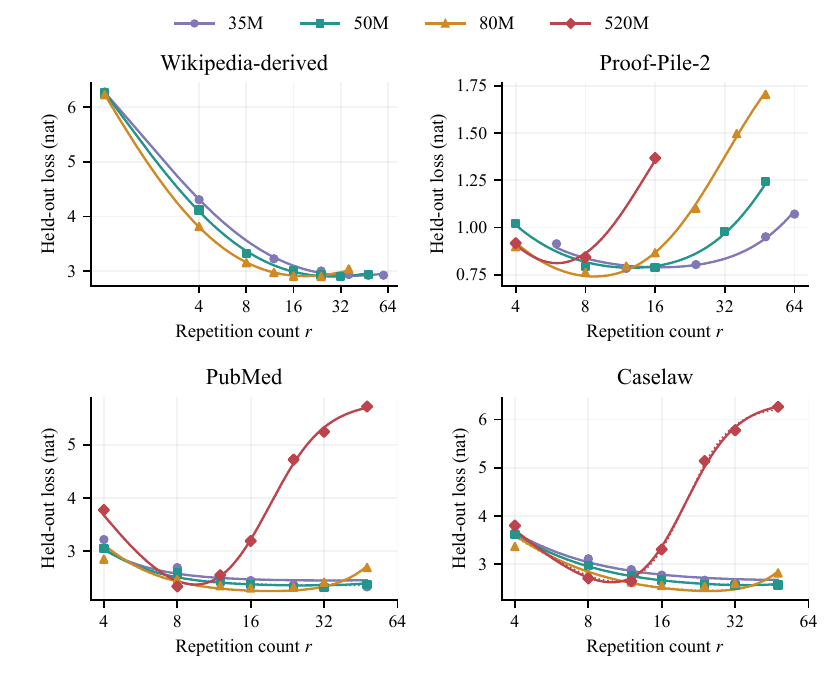}
 \caption{The same two-effect family fitted to all four corpora.
 Points are completed-run means. Solid lines show the central fit; dotted
 lines show the wider-bound sensitivity fit, both using Equation~\ref{eq:saturated-two-effect}.
 Agreement at the largest size here is reconstruction using those outcomes.}
 \label{fig:structural-all-curves}
\end{figure}

\subsubsection{An envelope generated by the structural law}

\begin{figure}[H]
 \centering
 \includegraphics[width=\textwidth]{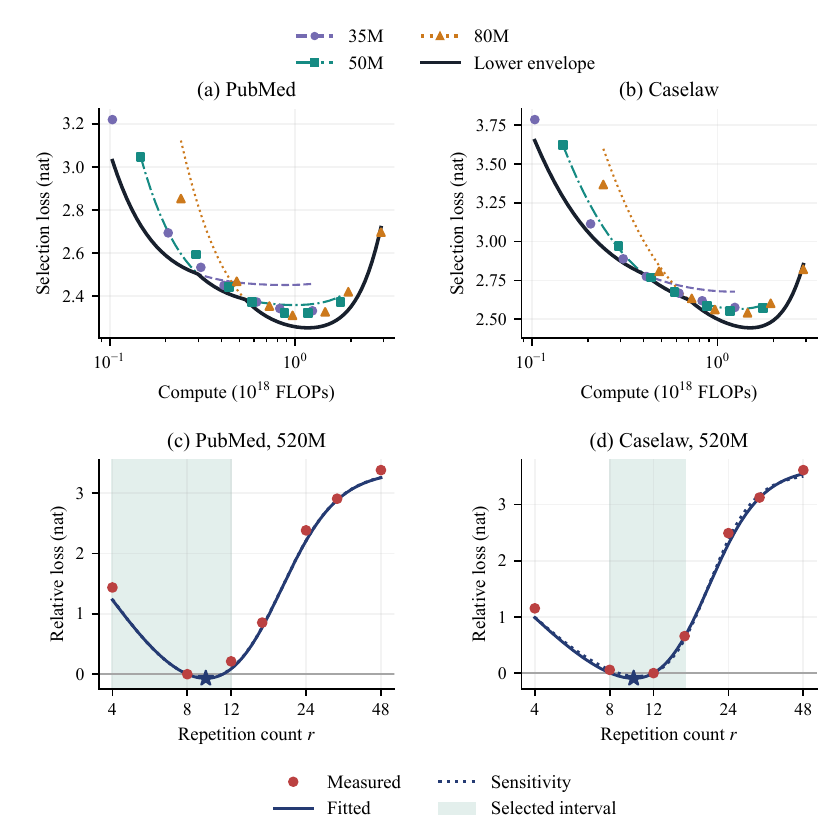}
 \caption{Compute envelopes and repetition-loss differences from the same empirical model.
 (a,b) Fits for 35M, 50M, and 80M on the common selection split; the black curve
 is their lower envelope within the tested repetition ranges, using $6ND$ as a compute proxy.
 (c,d) Independent 520M test losses relative to $r=8$ on PubMed and $r=12$ on Caselaw,
 compared with fitted differences anchored at the same respective counts.
 Stars mark untested continuous model minima; dotted curves show wider exponent limits.
 Shading spans the candidate intervals selected before 520M training.
 The curve model was fitted after the experiments.}
 \label{fig:structural-envelope-regret}
\end{figure}

For the fixed target pool and mixture fraction, let
$D(r)=125\mathrm M\,r$. The loss available at a compute value $C$
from one measured model size is
$\widehat L_N(C/(6N\,125\mathrm M))$.
We form the equal-compute envelope by taking the minimum over
$N\in\{35,50,80\}\mathrm M$, restricting each curve to its measured
repetition range. Figure~\ref{fig:structural-envelope-regret} shows this envelope together with the 520M loss differences.
The offsets are those estimated on the common smaller-model selection split.
The 520M independent test split is evaluated separately and is not pooled
into the absolute-loss envelope. The $6ND$ value is a compute proxy.

An equal-compute envelope can rise when larger repetition becomes harmful
within the available model family. Let $\widehat L_{=}(C)$ denote this minimum fitted loss at exact compute $C$.
If compute is an upper budget instead, the relevant envelope is $\min_{c\le C}\widehat L_{=}(c)$, which cannot
rise. Neither envelope searches unmeasured model families or hyperparameters.

\subsection{Additional numerical checks before the 200M study}\label{app:additional-comparisons}

{\color{ReviewPurple}
The following retrospective analyses compare the Lovelace form, estimate the scale exponent without fixing its sign, and predict omitted model sizes. Sensitivity of candidate pruning to smaller-model seeds and sizes is reported separately in Section~\ref{app:pruning-source-sensitivity}.
}

\subsubsection{Comparison with the additive repetition penalty}

\label{app:lovelace-comparison}

Equation (8) of \citet{lovelace2026prescriptivescalinglawsdata} is
\begin{equation}
 L(N,U_D,R_D)=E+\frac{A}{N^{\alpha_L}}
 +\frac{B}{[U_D(1+R_D)]^\beta}
 +P R_D^\delta\left(\frac{N}{U_D^\gamma}\right)^\kappa.
 \label{eq:app-lovelace-original}
\end{equation}
Here $U_D$ is the number of unique training tokens and $R_D$ counts
additional passes after the first. Thus the correspondence with our total
repetition count is $R_D=r-1$. Their experiments repeatedly train on a
single corpus, whereas ours repeat the target corpus while adding generic
tokens. To compare the repetition-dependent shapes at fixed target size,
we set $U_D=S_t$ and refit the coefficients, giving
\begin{equation}
 \widehat L_N(r)=A_N+b r^{-\beta}+pN^\kappa(r-1)^\delta.
 \label{eq:app-lovelace-shape}
\end{equation}
Using total rather than target training tokens in the decreasing term
only rescales $b$ at fixed $h$. We allow a separate offset $A_N$ at each
size, as in Section~\ref{app:structural-regret}. This comparison therefore
tests repetition-curve shape rather than the original model's absolute
loss scaling or its dependence on the unique-data budget. With one fixed
$S_t$, the original $P$ and $\gamma$ cannot be separately estimated.

Equation~\ref{eq:app-lovelace-shape} is a special case of
Equation~\ref{eq:two-effect-model}, with $\eta=0$, $q=\kappa$,
$U(r)=br^{-\beta}$, and $V(r)=p(r-1)^\delta$. It already permits a valley
and a change in the preferred repetition count with model size. The
additional question is whether allowing both terms to change in overall
amplitude improves their description of our mixture-training curves. We
first retain exactly the same repetition powers and only free $\eta$:
\begin{equation}
 \widehat L_N(r)=A_N+N^\eta
       \left[br^{-\beta}+pN^\kappa(r-1)^\delta\right].
 \label{eq:app-lovelace-nested}
\end{equation}
The two models have five and six shape parameters, respectively, in
addition to the offsets. We also compare them with the eight-parameter
saturating model already reported in Section~\ref{app:structural-regret}.
That extension changes several aspects of the model, so its full gain
cannot be attributed to $\eta$ alone.

\begin{table}[htbp]
 \centering
 \caption{Fits to all available pre-200M configuration means. Entries are
 RMSE in nats. The last column reproduces the existing central fit from
 Section~\ref{app:structural-regret}; all three columns use target-scale
 observations and describe fitted curves.}
 \label{tab:app-lovelace-fit}
 \normalsize
 \begin{tabularx}{\textwidth}{@{}l*{3}{>{\raggedleft\arraybackslash}X}@{}}
  \toprule
  Corpus & Lovelace shape & Only $\eta$ freed & Saturating extension \\
  \midrule
  Wikipedia-derived & 0.066 & 0.034 & 0.020 \\
  Proof-Pile-2 & 0.087 & 0.068 & 0.015 \\
  PubMed & 0.258 & 0.167 & 0.089 \\
  Caselaw & 0.263 & 0.202 & 0.076 \\
  \bottomrule
 \end{tabularx}
\end{table}

The comparison uses actual parameter counts and equally weights 97
configuration means: 21 for Wikipedia-derived text, 20 for Proof-Pile-2,
and 28 each for PubMed and Caselaw. For the two earlier corpora, the
smaller-model inputs are the archived reported means. PubMed and Caselaw
use the original selection-split means at 35M, 50M, and 80M and test means
at 520M. The later common-test reevaluation is not substituted into this
historical comparison.

We then omit each $r\in\{8,12,16,24,32\}$ at all four sizes for PubMed
and Caselaw, refit, and predict the omitted losses. Each corpus contributes
20 predictions. Every offset is estimated from the remaining counts at
its own size. Table~\ref{tab:app-lovelace-held-r} shows that the improvement
extends to repetition counts omitted from fitting.

\begin{table}[htbp]
 \centering
 \caption{RMSE on repetition counts omitted at every model size, in nats.
 The saturating-extension column is the earlier leave-count-out analysis
 from Section~\ref{app:structural-regret}, included as a reference.}
 \label{tab:app-lovelace-held-r}
 \normalsize
 \begin{tabularx}{\textwidth}{@{}l*{3}{>{\raggedleft\arraybackslash}X}@{}}
  \toprule
  Corpus & Lovelace shape & Only $\eta$ freed & Saturating extension \\
  \midrule
  PubMed & 0.336 & 0.236 & 0.107 \\
  Caselaw & 0.371 & 0.340 & 0.093 \\
  \bottomrule
 \end{tabularx}
\end{table}

These pre-200M Lovelace fits to all available sizes do not reach the exponent bounds;
raising their upper limit from 4 to 8 leaves the reported RMSE unchanged.
Some fits with free $\eta$ do reach a bound, so their individual exponents
are not precise parameter estimates. Replacing $(r-1)^\delta$ by $r^\delta$
while retaining free $\eta$ gives RMSEs of 0.030, 0.069, 0.168, and 0.202
for the four corpora in Table~\ref{tab:app-lovelace-fit}. This separate
check changes the repetition parameterization as well as freeing $\eta$.
The comparisons in this subsection constrain $\kappa$ to be positive;
the following subsection tests the sign independently. The evidence here
supports more accurate curve description and interpolation in our setting;
its implications for predicting an unseen model size are assessed below.

\subsubsection{Estimating the scale exponent without fixing its sign}

\label{app:q-sign-check}

We retain the repetition powers from Equation~\ref{eq:app-posthoc-fit} and
fit
\begin{equation}
 \widehat L_N(r)=A_N+n^\eta
   \left[B(r/8)^{-1/2}+Pn^q(r/8)\right],
 \qquad n=N/48{,}826{,}368.
 \label{eq:app-free-q}
\end{equation}
The coefficients $B,P$ remain nonnegative, while $q$ ranges over $[-8,8]$.
We fit both $\eta=0$ and $\eta\in[-4,4]$.
\newrev{We separately fit $q=0$ and obtain the $q\leq0$ comparator
from the nonpositive part of a 401-point profile over $[-8,8]$.}
Table~\ref{tab:app-q-sign} uses only the three smaller
sizes. The best fitted $q$ is positive in every corpus, with lower error
than the nonpositive alternative under the same choice of $\eta$.

\begin{table}[htbp]
 \centering
 \caption{Sign check using the 35M, 50M, and 80M curves. Error columns are
 RMSE in nats; ``$q\leq0$'' denotes the \newrev{best nonpositive profile value}.}
 \label{tab:app-q-sign}
 \normalsize
 \setlength{\tabcolsep}{4pt}
 \begin{tabular}{lrrrcrrr}
  \toprule
  & \multicolumn{3}{c}{$\eta=0$} && \multicolumn{3}{c}{Free $\eta$} \\
  \cmidrule(lr){2-4}\cmidrule(lr){6-8}
  Corpus & $q$ & RMSE & $q\leq0$ && $q$ & RMSE & $q\leq0$ \\
  \midrule
  Wikipedia-derived & 0.996 & 0.071 & 0.099 && 1.579 & 0.046 & 0.099 \\
  Proof-Pile-2 & 1.280 & 0.034 & 0.163 && 1.360 & 0.034 & 0.060 \\
  PubMed & 1.309 & 0.015 & 0.093 && 1.292 & 0.015 & 0.077 \\
  Caselaw & 1.175 & 0.021 & 0.088 && 1.170 & 0.021 & 0.067 \\
  \bottomrule
 \end{tabular}
\end{table}

We repeat the comparison with only 35M and 50M and with all available
sizes. All fitted signs remain positive. For $\eta=0$, the two-size
estimates are 1.016, 1.606, 1.261, and 0.927 in the same corpus order;
adding the 520M data changes the last three estimates to 0.633, 0.747,
and 0.864. Thus the direction is stable while the estimated exponent
depends on the fitted range. Profiles at 401 values of $q$ for each of
the 24 corpus/range/model combinations give 9,624 error records; all
best nonpositive profiles attain the $q=0$ boundary to numerical precision.
No best fit of Equation~\ref{eq:app-free-q} reaches its $q$ or $\eta$ bound.

For PubMed and Caselaw we resample complete seed curves within each
model size, independently across sizes. Applying the same sampled seed
identifiers to all counts preserves their within-curve dependence. Each
of the three fitting ranges uses 1,000 resamples with $\eta=0$ and 300
with free $\eta$, for 7,800 refits in total. Every refit has $q>0$.

\begin{table}[htbp]
 \centering
 \caption{Empirical 2.5th--97.5th percentile ranges of $q$ from resampling
 complete seed curves. ``All'' includes 520M. Each row uses 1,000 refits
 for $\eta=0$ and 300 for free $\eta$.}
 \label{tab:app-q-bootstrap}
 \normalsize
 \begin{tabular}{llcc}
  \toprule
  Corpus & Fitted sizes (M) & $\eta=0$ & Free $\eta$ \\
  \midrule
  PubMed & 35, 50 & [1.103, 1.384] & [0.894, 1.082] \\
  & 35, 50, 80 & [1.285, 1.334] & [1.263, 1.320] \\
  & All & [0.728, 0.772] & [0.647, 0.718] \\
  Caselaw & 35, 50 & [0.860, 0.998] & [0.688, 0.941] \\
  & 35, 50, 80 & [1.132, 1.213] & [1.123, 1.212] \\
  & All & [0.845, 0.887] & [0.819, 0.887] \\
  \bottomrule
 \end{tabular}
\end{table}

These ranges describe seed variability conditional on the fitted function
and training setup. Each smaller size has three observed seeds; the
number of resamples is not the number of independent training runs.
This additional analysis does not recover complete source-seed tables
for the two earlier corpora, so it reports no new bootstrap intervals
for them. The bootstrap in Section~\ref{app:posthoc-controls} fixes
$q=1$ and assesses selection stability, which is a different question.

We also retain the full function from
Equation~\ref{eq:saturated-two-effect} and only change its allowed
$q$ range from $[0.01,4]$ to $[-4,4]$. Twelve optimization starts include
both signs; separate fits impose $q\leq0$ and $q=0$. At $q=0$, $n_s$
is unidentifiable and is removed from the optimization. The best signs
are again positive (Table~\ref{tab:app-saturated-q}). Several estimates
reach the upper bound, supporting the direction more clearly than the
numerical value of the exponent. For Wikipedia-derived text fitted
only at 35M and 50M, the positive fit improves error by about 4\%; both
errors round to 0.016. This is the weakest comparison in the table.

\begin{table}[htbp]
 \centering
 \caption{Sign check in the original saturating function family. RMSEs
 are in nats; an asterisk marks the upper bound $q=4$. ``All'' means
 all pre-200M sizes available for that corpus.}
 \label{tab:app-saturated-q}
 \normalsize
 \begin{tabularx}{\textwidth}{@{}llrr>{\raggedleft\arraybackslash}X@{}}
  \toprule
  Corpus & Fitted sizes (M) & Best $q$ & RMSE & Best $q\leq0$ RMSE \\
  \midrule
  Wikipedia-derived & 35, 50 & 0.659 & 0.016 & 0.016 \\
  & 35, 50, 80 & 0.826 & 0.020 & 0.041 \\
  Proof-Pile-2 & 35, 50, 80 & $4.000^*$ & 0.012 & 0.055 \\
  & All & 3.523 & 0.015 & 0.065 \\
  PubMed & 35, 50, 80 & $4.000^*$ & 0.003 & 0.074 \\
  & All & $4.000^*$ & 0.089 & 0.215 \\
  Caselaw & 35, 50, 80 & $4.000^*$ & 0.004 & 0.058 \\
  & All & $4.000^*$ & 0.076 & 0.348 \\
  \bottomrule
 \end{tabularx}
\end{table}

\subsubsection{Predicting a model size omitted from fitting}

\label{app:held-scale-check}

The smaller-model fits also give a prediction for the best evaluated
count at a larger size. Table~\ref{tab:app-held-scale} tests the simple
specialization in Equation~\ref{eq:app-posthoc-fit}, keeping $q=1$ and
$\eta=0$. Fitting and selection use only the listed source sizes; target
losses are used to score the selected count. Because the target offset
is not predicted, these are checks of configuration choice and relative
loss, not of absolute loss at an unseen size. They are retrospective:
the function was developed after these experiments were available.

\begin{table}[htbp]
 \centering
 \caption{Retrospective held-size predictions from the simple model.
 Excess loss is the selected count's target mean loss minus the minimum
 over the target's evaluated counts, in nats.}
 \label{tab:app-held-scale}
 \normalsize
 \begin{tabularx}{\textwidth}{@{}l*{4}{>{\raggedleft\arraybackslash}X}@{}}
  \toprule
  Corpus & Source $\to$ target (M) & Predicted $r$ & Measured best $r$ & Excess loss \\
  \midrule
  Wikipedia-derived & $35,50\to80$ & 24 & 24 & 0 \\
  Proof-Pile-2 & $35,50\to80$ & 8 & 8 & 0 \\
  PubMed & $35,50\to80$ & 16 & 16 & 0 \\
  Caselaw & $35,50\to80$ & 24 & 24 & 0 \\
  \midrule
  Proof-Pile-2 & $35,50\to520$ & 4 & 8 & 0.074 \\
  PubMed & $35,50\to520$ & 4 & 8 & 1.435 \\
  Caselaw & $35,50\to520$ & 8 & 12 & 0.058 \\
  \bottomrule
 \end{tabularx}
\end{table}

The four 80M choices match the lowest measured means. The 520M results
are more variable: the PubMed choice has a substantial loss penalty,
whereas Caselaw selects a nearby low-loss count. Adding 80M to the
source fit leaves these three $q=1$ choices unchanged. Proof-Pile-2 has
only the three measured 520M candidates $\{4,8,16\}$.

Releasing $q$ preserves all four 80M choices, but does not consistently
improve the 520M predictions. Freeing both $q$ and $\eta$ changes the
Wikipedia-derived 80M choice to $r=16$ and the PubMed choice to $r=24$,
with excess losses of 0.008 and 0.016; the other two choices remain
unchanged. With 35M, 50M, and 80M sources, both the
free-$q$ model and the free-$(q,\eta)$ model select $r=4$ on all three
520M corpora; their excess losses are 0.074, 1.435, and 1.153.
Equations~\ref{eq:app-lovelace-shape} and
\ref{eq:app-lovelace-nested} also select $r=4$ in these three cases.
The saturating model fitted only to the three source sizes selects
$r=8,16,16$, with excess losses 0, 0.854, and 0.658, respectively.
On Wikipedia-derived text, its $35,50\to80$ prediction is $r=36$,
with excess loss 0.139. These comparisons separate improvements in
fitted or interpolated curves from reliable extrapolation across size.
The subsequent 200M experiment tests the predictions at an intermediate
scale before its results are used for fitting
(Section~\ref{app:intermediate-validation}).

\subsection{Predictions and empirical-model checks at 200M}\label{app:intermediate-model-checks}

{\color{ReviewPurple}
These checks use the completed 200M results in Section~\ref{app:intermediate-validation}. Predictions saved before training are evaluated separately from retrospective refits and additional-count diagnostics.
}

\subsubsection{Predicted minima, loss differences, and low-regret intervals}\label{app:200m-local-predictions}

The primary model is Equation~\ref{eq:app-posthoc-fit}. In the
normalization used by the frozen computation, its shape is
$g_N(r)=B(r/8)^{-1/2}+Pn(r/8)$, with
$n=N/48{,}826{,}368$. Its continuous minimum is
$8[B/(2Pn)]^{2/3}$. The saturated model instead uses
$n=N/(50\mathrm{M})$; the coefficients are interpreted with their
respective normalizations. The primary minima, 9.659 and 11.431,
lie 1.341 and 0.569 counts from the lowest measured means at 11 and 12.
These distances use measured grid minima as references, not known
continuous population optima.

Evaluating the unchanged primary function on integer counts selects
10 and 11, whose measured excess losses on $G_{12}$ are
$0.003\pm0.003$ and $0.008\pm0.003$ nats. The unchanged two-source
auxiliary function selects 11 and 12 on $G_{12}$. These integer-grid
checks use predictions fixed before training, but the finer-grid
evaluation and the detailed error summaries were added after the results.

To compare regret without fitting a target intercept, we subtract the
same reference count from each curve:
$\widehat\Delta(r;12)=\widehat L(r)-\widehat L(12)$ and
$\Delta(r;12)=L(r)-L(12)$. The exact regret expression in
Section~\ref{app:structural-regret} also covers the additive primary
model: with $u=\log(r/r_*)$ it reduces to
$g_N(r_*)[(2e^{-u/2}+e^u)/3-1]$.
Its local log-coordinate curvature is 0.835 for PubMed and 0.984 for
Caselaw. Subtracting two model regrets gives the predicted differences
in Table~\ref{tab:app200-loss-differences}.

\begin{table}[!htbp]
\centering
\caption{Frozen primary-model predictions and measured mean loss
differences relative to $r=12$, in nats. Values smaller in magnitude than
0.001 are shown as such to preserve their sign.}
\label{tab:app200-loss-differences}
\normalsize
\begin{tabular}{rrrrr}
\toprule
 & \multicolumn{2}{c}{PubMed} & \multicolumn{2}{c}{Caselaw} \\
$r$ & Predicted & Measured & Predicted & Measured \\
\midrule
8 & $-0.006$ & $0.038$ & $0.058$ & $0.095$ \\
10 & $-0.020$ & $-0.003$ & $0.007$ & $0.022$ \\
11 & $-0.013$ & $-0.006$ & $(-0.001,0)$ & $0.008$ \\
16 & $0.097$ & $0.100$ & $0.058$ & $0.073$ \\
48 & $1.575$ & $2.896$ & $1.425$ & $2.926$ \\
\bottomrule
\end{tabular}
\end{table}

The primary relative-loss RMSE is 0.021/0.019 nats for PubMed/Caselaw
on the seven measured counts between 8 and 16, but 0.560/0.641 over
the original seven counts spanning 4--48. The model underestimates the
large loss increase at high repetition. At the common center $r=11$,
the finite difference $L(10)-2L(11)+L(12)$ is predicted to be
0.007/0.008 and measured as $0.009\pm0.004$/$0.007\pm0.001$.
This checks local bending of the curve; it does not imply global convexity.

\begin{table}[!htbp]
\centering
\caption{The primary frozen model's interval with continuous model
regret at most 0.020 nats, compared with measured counts at most 0.020
above the $G_{12}$ minimum. Membership uses unrounded endpoints.
The continuous intervals are predictions, not confidence intervals.}
\label{tab:app200-regret-intervals}
\normalsize
\begin{tabularx}{\textwidth}{@{}l*{3}{>{\raggedright\arraybackslash}X}@{}}
\toprule
Corpus & Predicted interval & Measured counts inside & Measured low-loss counts \\
\midrule
PubMed & $[7.730,11.973]$ & $\{8,9,10,11\}$ & $\{9,10,11,12\}$ \\
Caselaw & $[9.314,13.936]$ & $\{10,11,12\}$ & $\{11,12,14\}$ \\
\bottomrule
\end{tabularx}
\end{table}

The interval estimates are near the low-loss regions but are shifted
left: they include PubMed $r=8$ and Caselaw $r=10$, which exceed the
measured threshold, and exclude PubMed $r=12$ and Caselaw $r=14$,
which meet it. No measurements certify the whole continuous interval.

\subsubsection{Checking the earlier empirical fit at the new scale}

The earlier fits in Section~\ref{app:structural-regret} used
35M/50M/80M selection losses and 520M test losses. Applying those saved
coefficients to 200M is a new-scale interpolation check, distinct from
the smaller-model-only predictions above. No 200M intercept was available,
so Table~\ref{tab:app200-old-fit} evaluates relative losses against the
same $r=12$ reference. ``Bound 4'' and ``bound 8'' identify the original
upper bounds on the fitted positive exponents.

\begin{table}[!htbp]
\centering
\caption{Unchanged earlier empirical coefficients evaluated at 200M.
The choice and its measured excess loss refer to $G_{12}$; the RMSE
columns compare $L(r)-L(12)$. The local grid is the seven measured counts
between 8 and 16.}
\label{tab:app200-old-fit}
\normalsize
\begin{tabularx}{\textwidth}{@{}llrr*{3}{>{\raggedleft\arraybackslash}X}@{}}
\toprule
Corpus & Earlier fit & $\widehat r_*$ & Choice & Excess loss & Full RMSE & Local RMSE \\
\midrule
PubMed & Bound 4 & 11.214 & 11 & 0 & 0.375 & 0.048 \\
 & Bound 8 & 9.525 & 10 & 0.003 & 0.598 & 0.171 \\
Caselaw & Bound 4 & 13.181 & 14 & 0.018 & 0.317 & 0.070 \\
 & Bound 8 & 11.639 & 12 & 0 & 0.598 & 0.068 \\
\bottomrule
\end{tabularx}
\end{table}

The bound-4 coefficients locate the PubMed minimum closely but predict
steeper valleys: their second differences at $r=11$ are 0.020/0.019,
against measured 0.009/0.007. Their 0.020-nat regret intervals are
$[9.825,12.699]$ and $[11.619,14.777]$. Every measured count inside
these two intervals meets the empirical threshold, although the intervals
miss the also-low-loss PubMed $r=9$ and Caselaw $r=11$. The Caselaw
bound-4 function prefers integer $r=13$, which was not measured; its
choice among the twelve measured counts is 14.

\subsubsection{Joint refits, held-out counts, and a fixed-size diagnostic}

The new joint fits retain Equation~\ref{eq:saturated-two-effect} and
use common-test means at five model sizes. There are 40 mean observations
per corpus: seven at each of 35M, 50M, 80M, and 520M, and twelve at 200M.
The default objective gives every mean observation equal weight and
profiles a separate intercept at each size. We distinguish fitting only
the original seven 200M counts from fitting all twelve. In the former,
the five additional counts enter neither the shape fit nor the 200M
intercept estimate.

\begin{table}[!htbp]
\centering
\caption{Checks of the same saturated empirical family. All errors
are in nats. Training errors, prediction errors, and relative-loss
errors are separated explicitly; the held-count rows use training-only
intercepts. The joint fits use exponent bound 4.}
\label{tab:app200-fit-diagnostics}
\normalsize
\begin{tabularx}{\textwidth}{Xrr}
\toprule
Check & PubMed & Caselaw \\
\midrule
Joint fit, all 40 means: training RMSE & 0.115 & 0.089 \\
Joint fit, all 12 target means: 200M training RMSE & 0.129 & 0.083 \\
Joint fit, seven target counts: absolute prediction RMSE on five added counts & 0.108 & 0.074 \\
Leave one count out at every size: RMSE over 30 predictions & 0.127 & 0.118 \\
Same leave-count-out fits: RMSE on ten 200M interior counts & 0.111 & 0.095 \\
Same leave-count-out fits: RMSE on five added 200M counts & 0.089 & 0.077 \\
\midrule
200M-only fit, seven counts: training RMSE & 0.002 & 0.001 \\
200M-only fit, seven counts: absolute prediction RMSE on five added counts & 0.004 & 0.006 \\
200M-only fit, seven counts: predicted continuous minimum & 10.805 & 11.855 \\
200M-only fit, all twelve counts: training RMSE & 0.002 & 0.003 \\
200M-only fit, all twelve counts: fitted continuous minimum & 10.786 & 11.819 \\
\bottomrule
\end{tabularx}
\end{table}

For the leave-count-out check, each of the ten interior counts in
$G_{12}$ is removed wherever it occurs, across all five sizes. The
model and all intercepts are refitted from the remaining observations
using fixed random initializations, without using the earlier full-data
coefficients. This produces 30 predictions per corpus: the five original
interior counts occur at five sizes and the five new counts only at 200M.

At a fixed $N$, the same empirical family has five effective shape
parameters: two coefficients, two repetition exponents, and the saturation
constant, plus one intercept. The 200M-only diagnostic estimates these
parameters without estimating $q$, $\eta$, or the scale-saturation location.
Its fit to the original seven counts predicts the five unused counts with
RMSE 0.004/0.006 and selects the measured minima, 11/12, on $G_{12}$.
The twelve-start fits reach the same best error without active parameter
bounds. This result supports the repetition-dependent function at the
tested size. The much larger joint-fit residuals suggest that the
shared cross-size parameterization is a remaining limitation, rather
than establishing which particular scale function should be changed.
Even the full single-size fit predicts a Caselaw second difference of
0.011 at $r=11$, above the measured 0.007.

\subsubsection{Matched comparison with the Lovelace form}

For the comparison in Table~\ref{tab:new-count-models}, all models use
exactly the same 35 training means per corpus: the original seven counts
at all five sizes. The five added 200M counts are excluded from both
parameter and intercept estimation. The fixed-target-pool adaptation of
Equation~(8) of \citet{lovelace2026prescriptivescalinglawsdata} uses the
shape $B(r/8)^{-\beta}+Pn^\kappa((r-1)/8)^\delta$ and a separate
intercept at each size. The nested comparison multiplies this same shape
by $n^\eta$, releasing only $\eta$. Thus the two models have five and
six fitted shape parameters, respectively. The saturated empirical model
has eight. Actual parameter counts are used in $n=N/(50\mathrm{M})$;
mean observations receive equal weight. The first two models use the
same eight optimization starts and exponent bounds.

The five-count absolute prediction RMSE decreases from 0.238 to 0.159
for PubMed and from 0.224 to 0.184 for Caselaw when $\eta$ is released,
reductions of 33.406\% and 18.156\%. The saturated model obtains
0.108/0.074, but changes additional structural assumptions.
Table~\ref{tab:app200-matched-point-errors} gives the individual residuals.
The improved prediction error does not imply improved selection:
the Lovelace and nested models both choose 10/11 on $G_{12}$, with
measured excess losses 0.003/0.008; the saturated joint model chooses
12/14, with 0.006/0.018. Releasing $\eta$ also increases the absolute
RMSE over all twelve 200M counts from 0.319/0.307 to 0.377/0.339.
The gain is therefore specific to prediction of these added counts.

\begin{table}[!htbp]
\centering
\caption{Prediction minus measured loss on the five 200M counts excluded
from the matched fits, in nats. Intercepts are estimated only from
training counts. ``$+\eta$'' changes only the scale-amplitude exponent;
``saturated'' is the joint empirical extension.}
\label{tab:app200-matched-point-errors}
\normalsize
\setlength{\tabcolsep}{5pt}
\begin{tabular}{rrrrrrr}
\toprule
 & \multicolumn{3}{c}{PubMed} & \multicolumn{3}{c}{Caselaw} \\
$r$ & Lovelace & $+\eta$ & Saturated & Lovelace & $+\eta$ & Saturated \\
\midrule
6 & 0.219 & 0.203 & 0.069 & 0.162 & 0.208 & 0.108 \\
9 & 0.226 & 0.103 & $-0.084$ & 0.206 & 0.148 & $-0.006$ \\
10 & 0.235 & 0.115 & $-0.105$ & 0.222 & 0.157 & $-0.032$ \\
11 & 0.242 & 0.134 & $-0.122$ & 0.236 & 0.169 & $-0.056$ \\
14 & 0.267 & 0.207 & $-0.144$ & 0.279 & 0.225 & $-0.106$ \\
\bottomrule
\end{tabular}
\end{table}

The PubMed Lovelace fit reaches the lower bound $\beta=0.100$; its
nested counterpart reaches $\delta=0.100$. Relaxing only these positive
exponent lower bounds to 0.010 gives added-count RMSE
0.239$\to$0.153 for PubMed and 0.224$\to$0.184 for Caselaw, preserving
the error ordering and every discrete choice. PubMed's baseline then has
$\beta=0.066$, away from the bound, while the nested fit still reaches
$\delta=0.010$. Its $\eta$ changes from 0.428 to 0.462. Caselaw has
no active bounds in either fit and $\eta=0.335$. The reused saturated
fits reach the upper bound $\nu=4$ for both corpora. Thus the added-count
gain survives this bounded sensitivity check, while individual exponents
remain insufficiently identified. These five counts belong to three
paired seed curves, not five independent experimental replications.

\subsubsection{Other sensitivity checks for the joint empirical fit}

Using common test losses without adding 200M changes the earlier bound-4
minima only from 11.214/13.181 to 11.220/13.203. Adding all twelve
200M counts gives 11.811/13.501. Keeping the earlier mixture of selection
and test splits while adding 200M instead gives 11.775/13.488, so
reevaluation alone does not explain the change.

With all five sizes, increasing the exponent upper bound from 4 to 8
gives minima 11.890/13.906 and twelve-point relative-loss RMSE
0.146/0.098, compared with 0.152/0.101 at bound 4. Giving each model
size equal total weight, rather than each mean observation, gives
11.889/13.697 and relative-loss RMSE 0.176/0.124. Removing one of the
three 200M seeds at a time, while keeping the source observations,
gives minimum ranges $[11.765,11.840]$ and $[13.460,13.532]$.
These are sensitivity ranges, not confidence intervals.

The remaining residuals also appear as a tradeoff across sizes.
Adding all 200M data reduces its centered shape RMSE from 0.284/0.238
to 0.129/0.083, but increases the corresponding 520M error from
0.070/0.072 to 0.156/0.149. The finer grid therefore supports accurate
fixed-size prediction in the chosen repetition-dependent family and
provides measurable gains for additional scale-amplitude freedom, while
also exposing limitations of the current shared cross-size fit.

\FloatBarrier
\section{Related Scaling Laws}\label{app:H}

\subsection{Where the decision sits}

The closest work can be organized by three questions: what is selected, what changes between inexpensive evidence and intended use, and whether candidate losses come from intermediate states of one training run or from independently trained runs. Kaplan et al. and Hoffmann et al. optimize model and token allocation \citep{kaplan2020scalinglawsneurallanguage,hoffmann2022trainingcomputeoptimallargelanguage}; repeated-data laws modify the value of processed tokens \citep{hernandez2022scalinglawsinterpretabilitylearning,xue2023repeatnottorepeat,muennighoff2025scalingdataconstrainedlanguagemodels,xu2026dataconstrainedlanguagemodelpretraining}; Liu et al. and related methods transfer data choices \citep{liu2025regmixdatamixtureregression,magnusson2025datadecidepredictbestpretraining,kang2025autoscalescaleawaredatamixing}; Sedova et al. and Zhou et al. explicitly model finite target reuse \citep{sedova2026scalinglawsmixturepretraining,zhou2026repetitionmismatchdatamixture}. Our decision is narrower: whether one evaluated repetition count preserves its ordering as model size changes.

\begin{table}[H]
\caption{Comparison with nearby scaling-law and data-selection studies. The third column states how evidence is used to choose a more expensive configuration.}
\label{tab:estimand_full}
\centering
\normalsize
\setlength{\tabcolsep}{3pt}
\begin{tabularx}{\linewidth}{@{}>{\raggedright\arraybackslash}p{0.25\linewidth}>{\raggedright\arraybackslash}p{0.20\linewidth}>{\raggedright\arraybackslash}p{0.24\linewidth}>{\raggedright\arraybackslash}X@{}}
\toprule
Work & Selected decision & From inexpensive evidence to intended use & Run construction and repetition \\
\midrule
\citet{kaplan2020scalinglawsneurallanguage}; \citet{hoffmann2022trainingcomputeoptimallargelanguage}
& Model size and token allocation & Model size, data, and compute & Processed tokens are treated as fresh \\

Repeated-data scaling \citep{hernandez2022scalinglawsinterpretabilitylearning,xue2023repeatnottorepeat,muennighoff2025scalingdataconstrainedlanguagemodels,xu2026dataconstrainedlanguagemodelpretraining}
& Useful epochs or finite-data allocation & Unique data, repeats, and model size & Repetition is explicit in an effective-data law \\

\citet{liu2025regmixdatamixtureregression}
& Data-mixture weights & Small proxy models to a larger run & Proxy regression over independently trained mixtures \\

\citet{sedova2026scalinglawsmixturepretraining}
& Mixture and repetition-aware minimum & Model size, token count, target size, and mixture & Candidate losses come from intermediate states of one run \\

\citet{zhou2026repetitionmismatchdatamixture}
& Mixture at a target token horizon & Token horizon and model size & Repetition is matched across training runs \\

DataDecide; AutoScale \citep{magnusson2025datadecidepredictbestpretraining,kang2025autoscalescaleawaredatamixing}
& Corpus ranking or domain weights & Model size or data scale & Completed corpus runs or domain-weight extrapolation \\

M$^3$ and multi-fidelity work \citep{akimoto2026m3scalinglawoptimizing,shukor2025scalinglawsoptimaldata,yen2025datamixtureoptimization}
& Training configuration or data mixture & Compute, data scarcity, or fidelity & Fitted response model or multi-scale evidence \\

\textbf{This work}
& \textbf{One evaluated repetition count} & \textbf{Separately trained smaller-model candidates to a larger model} & \textbf{One completed run per $(N,r,\mathrm{seed})$} \\
\bottomrule
\end{tabularx}
\end{table}

\paragraph{Model sizes and parameter accounting.}
Sedova et al. use GPT-2-style models spanning 101M--805M non-embedding parameters and report that larger models reach repetition-induced degradation earlier while attaining lower best-case loss \citep{sedova2026scalinglawsmixturepretraining}. This sub-billion-parameter range motivated the model sizes used in our study. However, our 34M--517M models use different architectures, and our parameter counts include embedding parameters \citep{sedova2026scalinglawsmixturepretraining}.

\subsection{From fresh-data scaling to constrained mixtures}

The parametric loss model of Hoffmann et al. decomposes loss into irreducible, finite-capacity, and finite-data terms \citep{hoffmann2022trainingcomputeoptimallargelanguage}:

\[
L(N,D)=E+\frac{A}{N^{\alpha}}+\frac{B}{D^{\beta}}.
\]

Here $N$ is parameter count, $D$ is the number of processed training tokens, $E$ is irreducible loss, and positive $A,B,\alpha,\beta$ set the finite-capacity and finite-data terms.

Its compute-optimal allocation does not distinguish a token's first appearance from a repeated exposure. In the single-corpus data-constrained setting, \citet{muennighoff2025scalingdataconstrainedlanguagemodels} instead write total tokens as $U_D(1+R_D)$, where $U_D$ is the number of tokens in the unique corpus and $R_D$ counts additional passes after the first, and replace $D$ by the following saturating effective amount of data:

\[
D'=U_D\left[1+R_D^*\left(1-e^{-R_D/R_D^*}\right)\right].
\]

Here $R_D^*>0$ is the fitted repetition scale at which additional passes begin to saturate. This recovers $D'\simeq U_D(1+R_D)=D$ for $R_D\ll R_D^*$ and approaches $U_D(1+R_D^*)$ as $R_D\to\infty$. The point here is that whole-corpus repetition changes the effective amount of data before the loss model of Hoffmann et al. is optimized.

Sedova et al. study a different training design: a constrained target corpus of size $D_{\mathrm{target}}$ is mixed with abundant generic data \citep{sedova2026scalinglawsmixturepretraining}. At target-token fraction $h$ and total token count $D_{\mathrm{total}}$, target repetition and its effective contribution are

\[
r=\frac{hD_{\mathrm{total}}}{D_{\mathrm{target}}},
\qquad
D_T=D_{\mathrm{target}}\left[1+r_1\left(1-e^{-(r-1)/r_1}\right)\right],
\]

where $r_1>0$ controls saturation and $\tau_{\mathrm{eff}}>0$ weights the target contribution. The resulting effective amount of training data is

\[
D_{\mathrm{eff}}=(1-h)D_{\mathrm{total}}+\tau_{\mathrm{eff}}D_T.
\]

Their fixed-size loss has the form $L_{\mathrm{fix}}=E+A/D_{\mathrm{eff}}^\alpha+\gamma h$, where $\gamma$ weights a target-fraction penalty; the multi-size form additionally includes a capacity term and an $N$-dependent data term. At fixed $D_{\mathrm{total}}$, increasing $h$ raises target exposure but reduces the generic-token allocation, so minimizing the law over $h$ can return an interior mixture optimum.

\subsection{How our independently completed configurations differ}

Our experiments hold $h$ and $D_{\mathrm{target}}$ fixed and train each repetition count to its own final token count. The design therefore satisfies $D_{\mathrm{total}}=rD_{\mathrm{target}}/h$. Substitution into the effective-data expression of Sedova et al. gives

\[
D_{\mathrm{eff}}(r)=D_{\mathrm{target}}
\left[
\frac{1-h}{h}r
+\tau_{\mathrm{eff}}\left(1+\rho(r)\right)
\right],
\qquad
\rho(r)=r_1\left(1-e^{-(r-1)/r_1}\right).
\]

For $0<h<1$ and $\tau_{\mathrm{eff}}>0$,

\[
\frac{\partial D_{\mathrm{eff}}}{\partial r}
=D_{\mathrm{target}}
\left[
\frac{1-h}{h}+\tau_{\mathrm{eff}}e^{-(r-1)/r_1}
\right]>0,
\]

so the effective amount of data increases with $r$. With positive loss coefficients and fixed $N,h$, this part of the loss decreases monotonically: the generic-token allocation continues to grow while the contribution from repeated target data saturates, and $\gamma h$ is constant. We are not testing the model of Sedova et al.; this calculation only shows that its effective-data term can motivate the decreasing component $U(r)$ in Equation~\ref{eq:two-effect-model} but cannot by itself produce the measured increase at high repetition. The increasing component $V(r)$ represents the additional behavior required by our loss curves.

\section{Secondary Wikipedia Diagnostics}\label{app:secondary-diagnostics}

\subsection{Secondary Wikipedia-Derived Diagnostics at a Glance}\label{app:G}

Table~\ref{tab:diagnostic-summary} summarizes what the secondary Wikipedia-derived analyses do and do not establish. These results test possible explanations and inexpensive diagnostic measurements; the main transfer claims rely on the directly measured loss curves in Sections~\ref{app:D} and~\ref{app:J}.

\begin{table}[H]
\centering
\caption{Questions addressed by the secondary Wikipedia-derived analyses.}
\label{tab:diagnostic-summary}
\normalsize
\setlength{\tabcolsep}{4pt}
\begin{tabularx}{\linewidth}{@{}>{\raggedright\arraybackslash}p{0.26\linewidth}>{\raggedright\arraybackslash}p{0.36\linewidth}>{\raggedright\arraybackslash}X@{}}
\toprule
Question & Observation & Supported conclusion \\
\midrule
Does a better 35M curve fit ensure transfer? & Conditions with small leave-one-out fit error still lose 0.032--0.035 nat at 50M. & Fit quality at one scale does not determine the transferred count (Section~\ref{app:B}). \\
Does gradient alignment predict the loss increase? & The fitted change is 0.001 nat per standard deviation, with an approximate 95\% interval of $[-0.002,0.005]$. & No large association is observed in this twelve-condition design; an exactly zero effect is not established (Section~\ref{app:A}). \\
Do simple text-overlap and duplication statistics predict the loss increase? & Lexical overlap is driven by two high-leverage conditions; longer $n$-gram overlap is nearly constant; {\color{ReviewPurple}8-gram redundancy is correlated with the corpus construction}. & None supplies a reliable prediction rule here (Section~\ref{app:A}). \\
Does changing the generic corpus remove the failure? & All twelve conditions select 48 at 35M and 32 at 50M, with {\color{ReviewPurple}0.032--0.042} nat loss increases from reuse. & The tested generic-corpus substitutions do not remove the transfer failure (Section~\ref{app:A}). \\
Do additional natural corpora extend the result? & Their gradient-alignment values cover the constructed range, but they have no loss curves across repetition counts. & They contextualize the measurement only (Section~\ref{app:F}). \\
Does higher repetition degrade performance on both corpus categories? & Held-out generic loss continues to improve in the matched-exposure comparisons. & The target-loss increase is not accompanied by an increase on the measured generic data (Section~\ref{app:matched-exposure-controls}). \\
\bottomrule
\end{tabularx}
\end{table}

Taken together, these analyses show that the observed Wikipedia-derived transfer failure persists across the tested generic-corpus changes and is not predicted by the measured low-cost diagnostics. They do not establish that all possible diagnostics must fail or that the same result holds for untested corpora.

\subsection{Statistical Status of the Secondary Wikipedia-Derived Analyses}\label{app:E}

The generic-corpus constructions, corpus measurements, regression form, and comparison between the two construction families in Section~\ref{app:A} were specified before the losses from the twelve conditions were evaluated. The accumulated-gradient estimator in Section~\ref{app:C} was also fixed before those training results were available. The reported regression interval uses a Student-$t$ distribution with six residual degrees of freedom.

The additional-corpus measurements in Section~\ref{app:F} and the simple transfer rules in Section~\ref{app:B} were evaluated after the corresponding loss curves were known. They are therefore presented as descriptive analyses and are not used to support the main repetition-transfer claims.

\subsection{Changing the Generic Training Corpus}\label{app:A}

This secondary Wikipedia-derived experiment asks whether the 35M-to-50M transfer failure disappears when the generic corpus in the training mixture changes. Across these constructed-corpus conditions, the target corpus, target-token fraction, token allocations, candidate sets, optimizer, and evaluation data remain fixed. This diagnostic uses $r\in\{6,12,24,48,64\}$ at 35M and $r\in\{4,8,16,32,48\}$ at 50M; these sets differ from those of the main Wikipedia-derived study.

\subsubsection{Generic-corpus constructions}

{\color{ReviewPurple}
We construct two families of generic corpora. The \emph{semantic-proximity} family uses \texttt{intfloat/e5-small-v2}, with the \texttt{passage:} prefix and normalized embeddings. We rank FineWeb documents by cosine similarity to the mean embedding of 20,000 target documents. The near set is the top 20\%; far candidates are the bottom 40\%. Each near document is matched to an unused far document using standardized log word count, word entropy, and within-document 10-gram redundancy. The pools mix near and far documents with near-set share $\kappa\in\{0,0.25,0.5,0.75,1\}$, using word counts to approximate the token allocation.

The \emph{entity-surface} family keeps the documents fixed but replaces a target-linked entity mention, with probability $\kappa$, by the written name or alias preferred in the target corpus; otherwise it uses a matched alias for the same entity. Before tokenization, both constructions remove a document when more than half its distinct lowercase word 8-grams occur in a target reference sampled up to 50M characters. This overlap filter is separate from the corpus statistics below.
}

Each family also has a duplication control at $\kappa=0$ in which 5\% duplicated examples are added. The ten main pools contain approximately 610--652M tokens after overlap removal. {\color{ReviewPurple}
The largest training configuration has a nominal budget of 1.6B generic tokens (a recorded maximum of 1.602B), so it can revisit these pools;
} {\color{ReviewPurple}
all conditions nevertheless use the same nominal generic-token budget at a fixed repetition count.
} The FineWeb generic corpus used in the main experiment is measured for comparison but is not one of the twelve constructed training conditions.

We use four seeds at $\kappa=0$ and $\kappa=1$ in each family, two at intermediate $\kappa$ values, and two for each duplication control, for 320 completed training runs (32 condition--seed combinations, two model sizes, and five counts per size) and 21 gradient-measurement runs. Table~\ref{tab:generic-properties} reports four corpus properties. Target--generic gradient alignment $m_c$ is the component of the preconditioned generic-data gradient in the target-gradient direction and is formally defined in Section~\ref{app:C}. {\color{ReviewPurple}
Lexical overlap is the fraction of distinct generic word bigrams that also occur in the target sample. Longer $n$-gram overlap uses the same ratio for word 8-grams. Within-generic 8-gram redundancy is one minus the number of distinct 8-grams divided by the total number of 8-gram occurrences. These statistics use the first 20,000 documents from each pool, whitespace-separated words with case preserved, and CRC32 hashes, with identical counting rules across conditions.
}

\begin{table}[H]
\centering
\caption{Measured properties of the twelve generic corpora. ``Semantic'' varies document proximity; ``entity'' varies target-preferred entity surface forms.}
\label{tab:generic-properties}
\normalsize
\setlength{\tabcolsep}{3pt}
\begin{tabularx}{\linewidth}{@{}>{\raggedright\arraybackslash}p{0.23\linewidth}r*{4}{>{\raggedleft\arraybackslash}X}r@{}}
\toprule
Condition & $\kappa$ & Gradient alignment & Lexical overlap & Longer $n$-gram overlap & {\color{ReviewPurple}
8-gram redundancy
} & Seeds \\
\midrule
Semantic 0.00 & 0.00 & 0.209 & 0.228 & 0.005 & 0.008 & 4 \\
Semantic 0.25 & 0.25 & 0.217 & 0.178 & 0.005 & 0.008 & 2 \\
Semantic 0.50 & 0.50 & 0.222 & 0.179 & 0.005 & 0.008 & 2 \\
Semantic 0.75 & 0.75 & 0.235 & 0.180 & 0.005 & 0.008 & 2 \\
Semantic 1.00 & 1.00 & 0.239 & 0.178 & 0.005 & 0.008 & 4 \\
Entity 0.00 & 0.00 & 0.393 & 0.189 & 0.005 & 0.007 & 4 \\
Entity 0.25 & 0.25 & 0.381 & 0.188 & 0.005 & 0.006 & 2 \\
Entity 0.50 & 0.50 & 0.365 & 0.188 & 0.005 & 0.005 & 2 \\
Entity 0.75 & 0.75 & 0.349 & 0.190 & 0.005 & 0.006 & 2 \\
Entity 1.00 & 1.00 & 0.335 & 0.194 & 0.005 & 0.006 & 4 \\
Semantic + 5\% duplicates & 0.00 & 0.208 & 0.231 & 0.005 & 0.057 & 2 \\
Entity + 5\% duplicates & 0.00 & 0.394 & 0.193 & 0.005 & 0.056 & 2 \\
\bottomrule
\end{tabularx}
\end{table}

{\color{ReviewPurple}
Within each family, $\kappa$ and gradient alignment are monotonically related. A test that permutes the five $\kappa$ labels separately within each estimation seed gives zero exceedances in 2,000 permutations for each family; Section~\ref{app:C} defines the statistic. Adding duplicates changes 8-gram redundancy by about 0.05 while changing gradient alignment by about 0.001. The main FineWeb corpus has alignment $0.222\pm0.007$, inside the semantic-proximity range.
}

\subsubsection{Repetition choices and transfer losses}

Every constructed corpus gives the same decision pattern: the lowest measured 35M loss occurs at $r=48$, and the lowest measured 50M loss occurs at $r=32$. Reusing 48 instead of 32 increases 50M loss by 0.032--0.042 nat. Table~\ref{tab:generic-transfer-results} reports the raw loss increase, which is the primary quantity.

\begin{table}[H]
\centering
\caption{Transfer results for the twelve generic corpora. All conditions select 48 at 35M and 32 at 50M.}
\label{tab:generic-transfer-results}
\normalsize
\begin{tabularx}{\linewidth}{@{}>{\raggedright\arraybackslash}p{0.29\linewidth}r*{3}{>{\raggedleft\arraybackslash}X}@{}}
\toprule
Condition & Seeds & 35M selection & 50M selection & Loss increase from reusing 48 \\
\midrule
Semantic 0.00 & 4 & 48 & 32 & 0.042 \\
Semantic 0.25 & 2 & 48 & 32 & 0.034 \\
Semantic 0.50 & 2 & 48 & 32 & 0.033 \\
Semantic 0.75 & 2 & 48 & 32 & 0.032 \\
Semantic 1.00 & 4 & 48 & 32 & 0.037 \\
Entity 0.00 & 4 & 48 & 32 & 0.036 \\
Entity 0.25 & 2 & 48 & 32 & 0.036 \\
Entity 0.50 & 2 & 48 & 32 & 0.035 \\
Entity 0.75 & 2 & 48 & 32 & 0.039 \\
Entity 1.00 & 4 & 48 & 32 & 0.037 \\
Semantic + 5\% duplicates & 2 & 48 & 32 & 0.037 \\
Entity + 5\% duplicates & 2 & 48 & 32 & 0.035 \\
\bottomrule
\end{tabularx}
\end{table}

{\color{ReviewPurple}
All available paired differences are positive; for Semantic 0.00 they are $\{0.043,0.037,0.040,0.051\}$ nat. Their mean is computed before rounding and rounds to 0.042 nat.
} Moving from 32 to 48 also adds 2B training tokens under Equation~\ref{eq:total-tokens}, so the longer configuration is both more expensive and worse. Changing the tested generic corpus therefore does not remove the transfer failure.

The smaller-model curve fit is not informative about this result. Only three of the twelve effective-data fits have leave-one-out root-mean-square error below $6.4\times10^{-3}$ nat, but those three still have transfer losses of 0.032--0.035 nat. A quadratic in $\log r$, used as a more flexible three-parameter comparison, also maps to 48 at 50M in all twelve conditions. Section~\ref{app:B} gives the fitted baseline and other simple transfer rules.

\subsubsection{Association with gradient alignment, text overlap, and duplication}

We next ask whether the size of the transfer loss varies with the measured corpus properties. For condition $i$, let $y_i$ be the 35M-to-50M loss increase. We fit the seed-count-weighted regression
\[
y_i=\alpha+\beta_\kappa\kappa_i+\beta_d d_i+\beta_n n_i+\beta_l l_i+\beta_c m_{c,i}+\epsilon_i,
\]
where $\alpha$ is the intercept, $d_i$, $n_i$, and $l_i$ are {\color{ReviewPurple}
8-gram redundancy
}, longer $n$-gram overlap, and lexical overlap, and $\epsilon_i$ is the residual. The fit has 12 observations and 6 residual degrees of freedom.

\begin{table}[H]
\centering
\caption{Weighted regression of transfer loss on the constructed corpus properties.}
\label{tab:generic-regression}
\normalsize
\begin{tabular}{lrrr}
\toprule
Term & Coefficient & Standard error & $t$ statistic \\
\midrule
Intercept & -0.135 & 0.135 & -1.00 \\
Construction strength $\kappa$ & 0.001 & 0.002 & 0.75 \\
Gradient alignment $m_c$ & 0.019 & 0.016 & 1.18 \\
Lexical overlap & 0.233 & 0.088 & 2.64 \\
Longer $n$-gram overlap & 24.54 & 23.31 & 1.05 \\
{\color{ReviewPurple}
8-gram redundancy
} & -0.078 & 0.037 & -2.11 \\
\bottomrule
\end{tabular}
\end{table}

One standard deviation of gradient alignment is 0.076, so its fitted association corresponds to only 0.001 nat of transfer loss. A two-sided 95\% Student-$t$ interval computed from Table~\ref{tab:generic-regression} is approximately $[-0.002,0.005]$ nat per standard deviation. The point estimate is therefore small and uncertain; it does not show that the true association is exactly zero.

The lexical and duplication coefficients are also descriptive. Ten conditions occupy a narrow lexical-overlap range of 0.178--0.194, while two conditions lie near 0.23, so the lexical estimate is driven by two high-leverage points. Longer $n$-gram overlap is nearly constant. Within-family fits give gradient-alignment slopes of 0.056 (standard error 0.065) for the semantic construction and 0.001 (standard error 0.027) for the entity construction. None of these measurements provides a reliable rule for predicting the transfer loss in a new corpus.

\begin{figure}[H]
\centering
\includegraphics[width=0.6\linewidth]{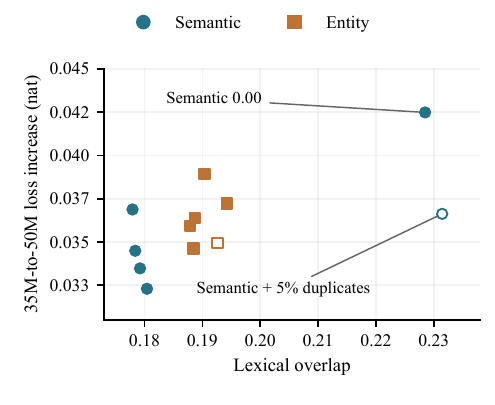}
\caption{The descriptive lexical-overlap association is driven by two conditions outside the narrow range occupied by the other ten. Labels identify the two semantic conditions with unusually high overlap; hollow markers denote the duplication controls. The vertical axis uses the same nat loss differences as Table~\ref{tab:generic-transfer-results}. The figure is a sensitivity diagnostic, not evidence for a lexical mechanism.}
\label{fig:A1}
\end{figure}

Section~\ref{app:C} describes the gradient measurement, and Section~\ref{app:E} distinguishes the prespecified analyses from the later descriptive analyses.

\subsection{Simple Rules Applied to the Wikipedia-Derived Loss Curves}\label{app:B}

This section asks whether the measured 35M Wikipedia-derived curve suggests a simple rule that performs well at both 50M and 80M. These rules were evaluated after the loss curves were available. They are baselines, not new methods or independent tests.

\subsubsection{Rules}

We consider fixed counts, direct reuse of the 35M selection, a single-corpus effective-data fit, and a tokens-per-parameter rescaling. The fitted baseline uses
\[
L(r)=E+B[D_{\mathrm{eff}}(r)]^{-\beta},\qquad
D_{\mathrm{eff}}(r)=S_t\left[1+R^*\left(1-e^{-(r-1)/R^*}\right)\right],
\]
Here $D_{\mathrm{eff}}(r)$ is the repetition-adjusted effective token count, $R^*=10$ fixes its saturation scale, $E$ is an additive loss floor, $B$ is a positive scale coefficient, and $\beta$ is the fitted decay exponent. We fit $E,B,\beta$ to the 35M curve. This is the single-corpus repeated-data form reviewed in Appendix~\ref{app:H}, applied here only as a baseline at fixed model size. The tokens-per-parameter rule scales the 35M count in proportion to the realized parameter counts in Appendix~\ref{app:I} before mapping to the available candidate set.

\subsubsection{Results}

\begin{table}[H]
\centering
\caption{Simple rules applied to the measured Wikipedia-derived curves. Loss increase is relative to the lowest measured mean at each scale: $r=32$ at 50M and $r=24$ at 80M.}
\label{tab:wikipedia-rules}
\normalsize
\setlength{\tabcolsep}{4pt}
\begin{tabularx}{\linewidth}{@{}>{\raggedright\arraybackslash}p{0.32\linewidth}*{4}{>{\raggedleft\arraybackslash}X}@{}}
\toprule
Rule & Count at 50M & Loss increase & Count at 80M & Loss increase \\
\midrule
Fixed $r=4$ & 4 & 1.208 & 4 & 0.914 \\
Fixed $r=8$ & 8 & 0.419 & 8 & 0.251 \\
Fixed $r=16$ & 16 & 0.104 & 16 & 0.008 \\
Fixed $r=24$ & 24 & 0.020 & 24 & 0 \\
Fixed $r=32$ & 32 & 0 & 36$^\dagger$ & 0.139 \\
Reuse 35M selection & 48 & 0.033 & 36 & 0.139 \\
35M effective-data fit & 48 & 0.033 & 36 & 0.139 \\
Tokens-per-parameter rescaling & 48 & 0.033 & 36 & 0.139 \\
\bottomrule
\end{tabularx}
\end{table}

$^\dagger$ Count 32 is absent from the 80M candidate set and maps to 36 under the log-distance rule in Section~\ref{sec:method}. Count 16 is close to the 80M minimum but is much worse at 50M. Conversely, the lowest measured mean at either larger scale does not transfer to the other. No listed rule attains the lowest measured mean at both scales.

The three rules based on the 35M curve produce the same evaluated choices for different reasons. The effective-data fit is monotone decreasing, so minimizing it over the complete 35M candidate set selects 60. The log-distance mapping sends 60 to 48 at 50M and 36 at 80M. Direct reuse starts from the empirical 35M choice of 48 and reaches the same two evaluated counts. Tokens-per-parameter rescaling recommends continuous counts of approximately 68 at 50M and 112 at 80M, both beyond the available candidate sets, so it also maps to the high-repetition boundary.

\subsubsection{Fit quality does not determine the transferred decision}

The same effective-data family was applied to the twelve generic-corpus conditions in Section~\ref{app:A}. For this diagnostic, $R^*=10$ and $\beta=0.5$ are fixed and $E,B$ are fitted. The reported leave-one-out errors use the four source counts $r\in\{12,24,48,64\}$; errors over the full five-count source range are larger in every condition. Minimizing the decreasing fitted curve over the source set selects 64, which maps to the largest evaluated 50M count, 48. Three conditions have leave-one-out root-mean-square error below $6.4\times10^{-3}$ nat, yet all three still transfer $r=48$ and incur 0.032--0.035 nat at 50M. Across all twelve conditions, fit error and transfer loss have Spearman correlation $\rho=0.56$, while every fitted decision is the same.

\begin{figure}[H]
\centering
\includegraphics[width=0.8\linewidth]{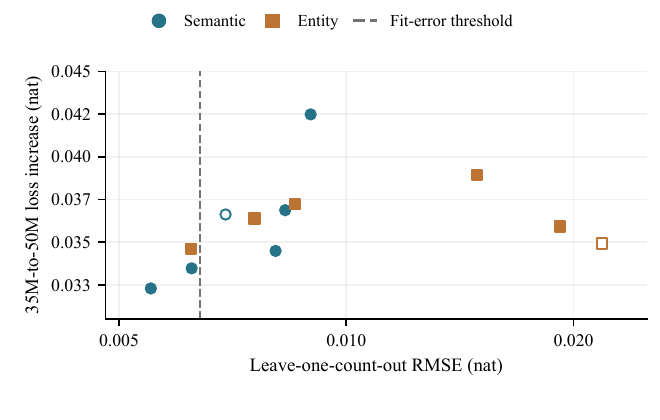}
\caption{Leave-one-out fit error at 35M versus the measured 35M-to-50M loss increase. Conditions with the smallest fit errors still transfer $r=48$, which has higher 50M loss than $r=32$. The dashed line marks the $6.4\times10^{-3}$-nat fit-error threshold; circles and squares distinguish the semantic and entity families, with hollow markers for duplication controls. The plot concerns this fitted baseline; it is not a claim that every possible scaling model must fail.}
\label{fig:fitscatter}
\end{figure}

These baselines show why a good fit to one smaller-model loss curve is not by itself a transfer guarantee. They do not rule out a method that explicitly models how candidate losses change with model size.

\subsection{Measuring Target--Generic Gradient Alignment}\label{app:C}

Section~\ref{app:A} uses one diagnostic for how strongly a generic corpus points in the same update direction as the Wikipedia-derived target. In plain terms, $m_c$ is the component of the generic-data gradient that points in the target-gradient direction after Adam preconditioning. It is used only as a descriptive corpus measurement, not as a mechanism or an input to the repetition decision.

\subsubsection{Definition}

The \emph{reference model state} is a saved set of model parameters and Adam optimizer statistics at a specified point in a 35M training run. Let $\mu_T$ and $\mu_G$ be population gradients for the target and generic data at this fixed state. Let $M=\mathrm{diag}(v)^{-1/2}$ be the Adam second-moment preconditioner stored with it. We define
\[
m_c=\frac{\langle M\mu_T,M\mu_G\rangle}{\lVert M\mu_T\rVert^2}.
\]
Positive values mean that part of the generic update aligns with the target update. Because the scale depends on both the model state and preconditioner, absolute values should be compared only under the same reference.

We estimate the gradients by independently accumulating $K$ target minibatches and $K$ generic minibatches. The target gradient appears in both the numerator and denominator, so minibatch noise attenuates the direct estimate. Let $\mathrm{self}(K)$ denote the expected alignment between two independent target-gradient estimates, each accumulated over $K$ minibatches. Under an independent additive-noise approximation, $\mathrm{self}(K)\approx1/(1+c/K)$, where $c$ is the preconditioned noise-to-signal ratio.

\subsubsection{Choosing the accumulation depth}

\begin{table}[H]
\centering
\caption{Gradient-alignment estimates at different accumulation depths for one saved 35M model state. Target-to-target uses independent target minibatches; FineWeb-to-target uses the main generic corpus.}
\label{tab:gradient-accumulation}
\normalsize
\begin{tabular}{lrrrrr}
\toprule
$K$ minibatches & 1 & 8 & 32 & 64 & 128 \\
\midrule
Target-to-target alignment & 0.059 & 0.320 & 0.619 & 0.770 & 0.868 \\
Unpreconditioned cosine & 0.41 & 0.85 & 0.96 & 0.98 & 0.99 \\
FineWeb-to-target alignment & 0.025 & 0.124 & 0.229 & 0.278 & 0.298 \\
\bottomrule
\end{tabular}
\end{table}

{\color{ReviewPurple}
The fitted reliability curve has $c\approx19.5$ and agrees closely with the measured target-to-target value at $K=128$. We use $K=32$, six independent estimation draws, and five estimation seeds. For each family, the permutation test first averages the six draws within each seed and condition. Its statistic is the range of the five condition means across seeds. We independently permute the five $\kappa$ labels within each seed, recalculate this range, and count values at least as large as observed over 2,000 permutations. This preserves the estimation-seed blocks and tests variation across $\kappa$ at a fixed reference state.
}

\begin{figure}[H]
\centering
\includegraphics[width=0.6\linewidth]{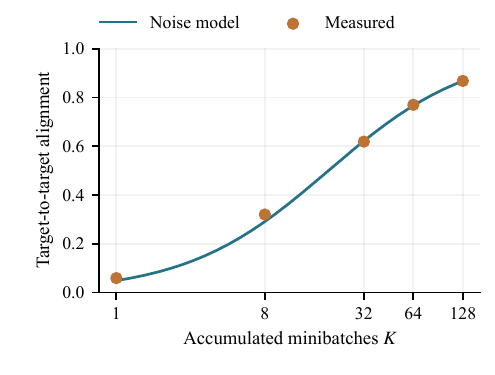}
\caption{Measured target-to-target gradient alignment as accumulation depth increases, together with $1/(1+19.5/K)$. This checks measurement noise; it does not test whether a count selected at one model size remains preferable at another.}
\label{fig:C1}
\end{figure}

\subsubsection{Dependence on the reference model state}

{\color{ReviewPurple}
Comparisons across four saved reference states preserve the direction of both constructed corpus changes, although the absolute values shift and can become negative. At a separate 35M reference state trained only on FineWeb ($h=0$), the semantic-proximity response has range 0.058, with zero exceedances in 2,000 within-seed permutations. The entity-surface response has range 0.008, below its mean across-seed standard deviation of 0.018; 130 of 2,000 permutations exceed its observed range ($p=0.065$). This comparison uses the smaller calibration pools for the ten main conditions. All measurements in Section~\ref{app:A} use the full pools and the same reference state trained with target-token fraction $h=0.8$. Gradient alignment therefore depends on both the corpus and the reference state.
}

Across the twelve constructed conditions, {\color{ReviewPurple}
8-gram redundancy
} and lexical overlap explain 77\% of the observed variation in $m_c$, mostly through bigram overlap. Section~\ref{app:A} therefore includes these text-overlap and duplication measurements in the regression. The result concerns this pretraining-mixture experiment and does not evaluate gradient-based data-selection methods such as TracIn, LESS, or DoGE \citep{pruthi2020estimatingtrainingdatainfluence,xia2024lessselectinginfluentialdata,fan2024dogedomainreweightinggeneralization}.

\subsection{Gradient Alignment on Additional Corpora}\label{app:F}

This descriptive analysis applies the estimator from Section~\ref{app:C} to additional corpora. It provides scale and coverage for the diagnostic only: no repetition-count sweep was trained on these corpora, so the measurements do not extend the transfer result.

\begin{table}[H]
\centering
\caption{Target--generic gradient alignment at the fixed Wikipedia-derived reference model state defined in Section~\ref{app:C}, using $K=32$ and five estimation seeds. ``Seed SD'' denotes the standard deviation across estimation seeds.}
\label{tab:natural-gradient-alignment}
\normalsize
\begin{tabularx}{\linewidth}{@{}>{\raggedright\arraybackslash}X>{\raggedleft\arraybackslash}p{0.16\linewidth}r>{\raggedright\arraybackslash}X@{}}
\toprule
Corpus & Mean alignment & Seed SD & Relation to the constructed ranges \\
\midrule
Project Gutenberg & 0.105 & 0.005 & Below semantic range \\
OpenAssistant-1 & 0.184 & 0.003 & Below semantic range \\
FineWeb generic & 0.222 & 0.007 & Inside semantic range \\
Unmodified base corpus & 0.232 & 0.004 & Near semantic range \\
Harvard USPTO Patent Dataset (HUPD) claims & 0.271 & 0.013 & Between constructed ranges \\
Raw Wikipedia & 0.307 & 0.004 & Between constructed ranges \\
Small news-text sample & 0.308 & 0.008 & Between constructed ranges \\
HUPD patent prose & 0.341 & 0.024 & Inside entity range \\
TriviaQA & 0.436 & 0.007 & Above entity range \\
Proof-Pile-2 formal-mathematics/code sample & 0.471 & 0.010 & Above entity range \\
Wikipedia-derived target, independent minibatches & 0.624 & 0.005 & Target-to-target reference \\
\bottomrule
\end{tabularx}
\end{table}

The unmodified base corpus uses the same source documents as the entity-surface family before any alias replacement. It differs from the $\kappa=0$ condition in Section~\ref{app:A}, which replaces target-linked mentions with matched, non-target-preferred aliases. Project Gutenberg, OpenAssistant-1, the news sample, and TriviaQA each contain fewer than 10M tokens, so sampled windows can recur across estimation draws. The additional corpora fill the gap between the semantic-proximity range $[0.208,0.239]$ and the entity-surface range $[0.335,0.394]$. This shows that the diagnostic is not confined to two artificial bands. It does not show how repetition transfer behaves in the intervening corpora; answering that question would require new training experiments.

\end{document}